\RequirePackage{fix-cm}
\documentclass[twocolumn]{svjour3}          % twocolumn
\usepackage{graphicx}
\usepackage[T1]{fontenc}
\usepackage[latin9]{inputenc}
\usepackage{color}
\usepackage{array}
\usepackage{float}
\usepackage{booktabs}
\usepackage{mathrsfs}
\usepackage{multirow}
\usepackage[ruled]{algorithm2e}
\usepackage{amsmath}
\usepackage{amssymb}
\usepackage{graphicx}
\usepackage{esint}
\makeatletter

\providecommand{\tabularnewline}{\\}

\usepackage{subfigure}
\usepackage{cite}
\usepackage{blindtext}
\usepackage{algorithmic}
\usepackage{graphicx}
\usepackage[export]{adjustbox}
\usepackage{hyperref}
\makeatother

\usepackage[english]{babel}
\begin{document}
\sloppy
\title{Shape Estimation for Elongated Deformable Object using B-spline Chained
Multiple Random Matrices Model%\thanks{Grants or other notes
}

\author{Gang Yao \and Ryan Saltus \and  Ashwin Dani}

\date{Received: date / Accepted: date}
\institute{Gang~Yao \at 
              \email{gang.yao@uconn.edu}           %  \\
           \and
           Ryan~Saltus \at
            \email{ryan.saltus@uconn.edu}  
           \and
          Ashwin~Dani \at 
           \email{ashwin.dani@uconn.edu} 
        \at
        \at Department of Electrical and Computer Engineering, University of Connecticut, Storrs, Connecticut, USA.\\}

\maketitle
\begin{abstract}
In this paper, a B-spline chained multiple random
matrices representation is proposed to model geometric characteristics
of an elongated deformable object. The hyper degrees
of freedom structure of the elongated deformable object make its shape
estimation challenging. Based on the likelihood function of the proposed
model, an expectation-maximization (EM) method is derived to estimate
the shape of the elongated deformable object. A split
and merge method based on the Euclidean minimum spanning tree (EMST)
is proposed to provide initialization for the EM algorithm. The proposed
algorithm is evaluated for the shape estimation of the
elongated deformable objects in scenarios, such as the static rope
with various configurations (including configurations with intersection), the continuous manipulation of a rope
and a plastic tube, and the assembly of two plastic tubes. The execution
time is computed and the accuracy of the shape estimation results
is evaluated based on the comparisons between the estimated width
values and its ground-truth, and the intersection over union (IoU)
metric. 
\end{abstract}

\section{Introduction}
\label{intro}

Elongated deformable objects are deformable objects which are characterized
by a length that is much longer than their width \cite{shah2018planning,zea2016tracking}.
Objects of this type are commonly encountered in daily life, such
as ropes, tubes, and trains. Automatic manipulation tasks such as
grasping, completing surgical sutures or assembling cable harnesses
are challenging due to the hyper degrees of freedom structure of the
elongated deformable objects \cite{sanchez2018robotic,shah2018planning,jackson2017real}.
To improve the manipulation performance, it is necessary to provide
an accurate perception model of the elongated deformable object as
feedback to the robotic manipulators. The shape estimation of
the elongated deformable object using data collected from the perception
sensors, such as RGB-D cameras, is a challenging problem. In this
paper, a shape estimation methodology for elongated deformable objects
using a chained multiple random matrices representation is developed.

\begin{figure}
\begin{centering}
\makebox[0.7\linewidth][c]{\subfigure[]{\includegraphics[width=0.87\columnwidth,fbox]{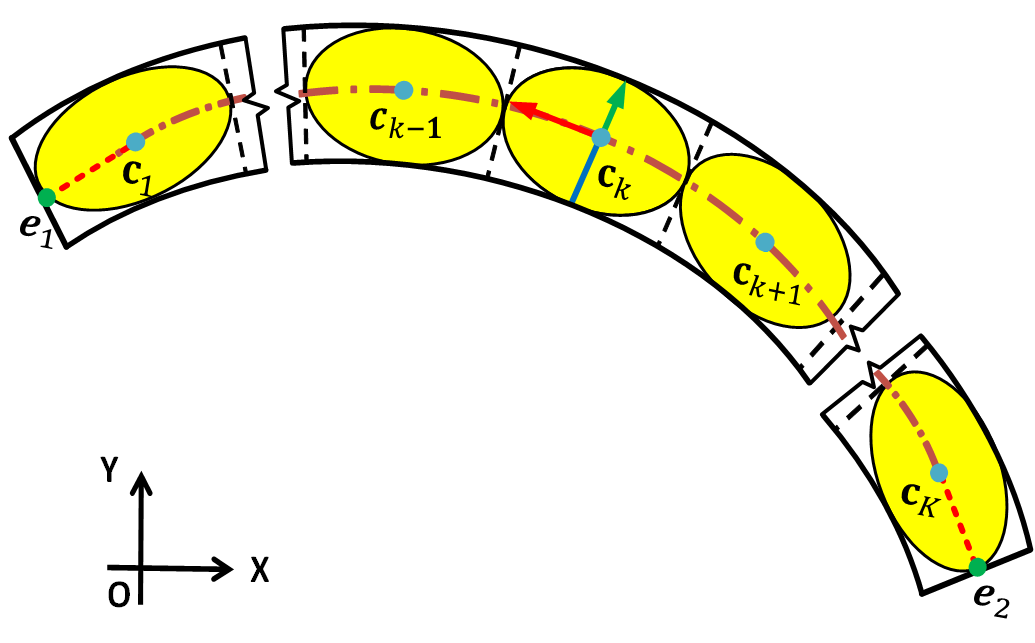}}}\\
\par\end{centering}
\begin{centering}
\makebox[0.7\linewidth][c]{\subfigure[]{\includegraphics[width=0.4\columnwidth,height=3cm,fbox]{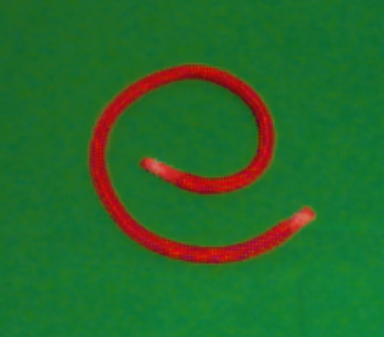}}\hspace{1em}\subfigure[]{\includegraphics[viewport=0bp 0bp 2608bp 2583bp,width=0.4\columnwidth,height=3cm,fbox]{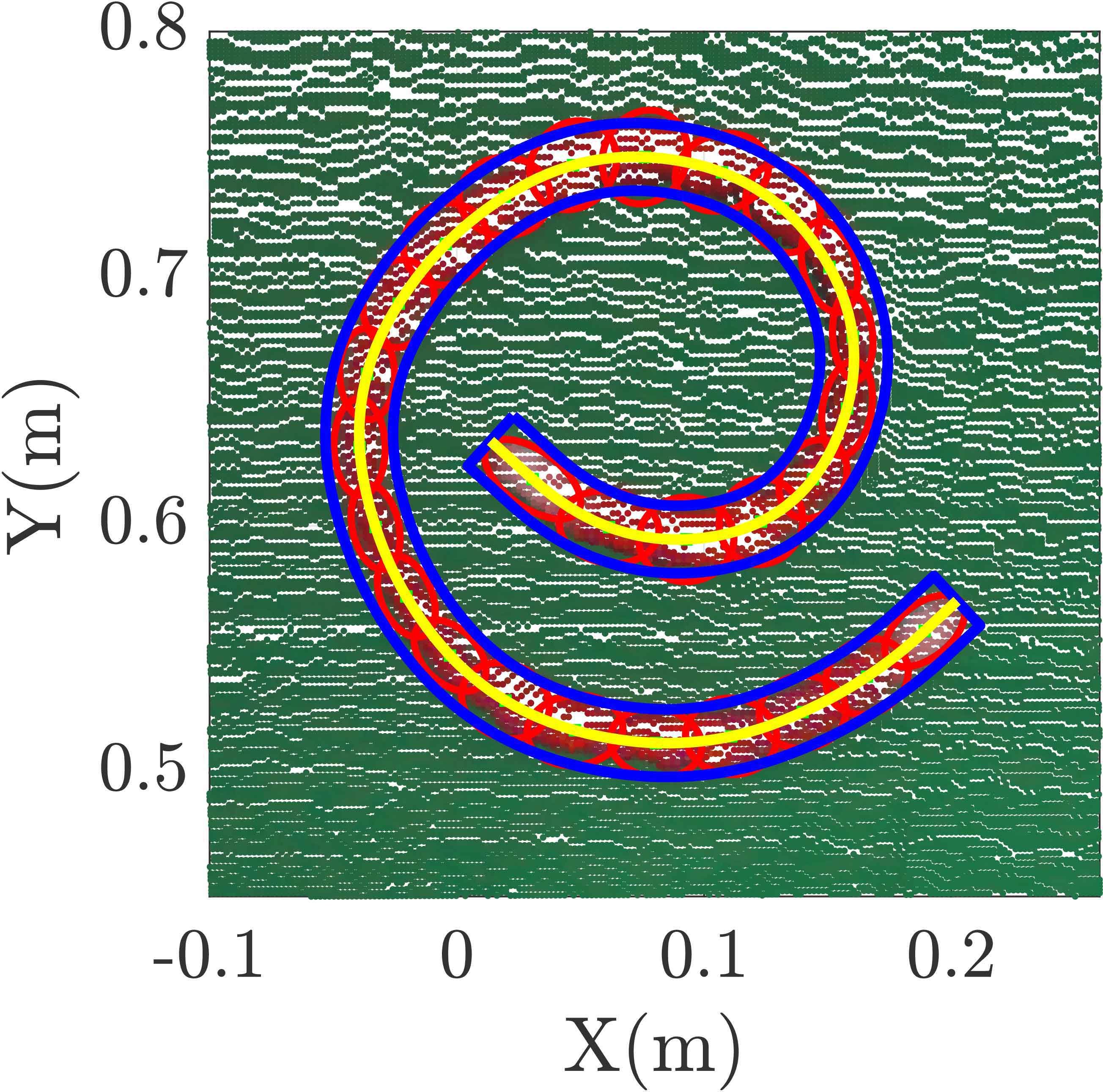}}}\\
\par\end{centering}
\centering{}\caption{(a) Representation of an elongated deformable object as a B-spline
curve chained multiple random matrices model; (b) The RGB image of
the rope; (c) The point cloud and the shape estimation result of the
rope, where the red ellipses are chained by the yellow B-spline curve
and the blue curves are the offset curves from the B-spline. \label{fig:Rope as MM}}
\end{figure}

Some works use the physics-based simulation models such as linked capsules \cite{tang2018track,schulman2013tracking}, mass-spring models or finite element method model  \cite{petit2015real} to represent the elongated deformable object
by considering the physical constraints of the object \cite{shah2018planning,tang2018track,petit2015real,schulman2013tracking,javdani2011modeling}.
A variety of registration algorithms are used to find the correspondences
between the measurements and the predefined nodes on the simulated
physics model. For example, the Gaussian mixture model (GMM) incorporating
coherent point drift regularization is applied to register the rope
nodes of a dynamic simulation model to the noisy point cloud, using
a stereo camera in \cite{tang2018track}. The iterative closest point
(ICP) method is used to estimate the rigid transformation from a point
cloud to a 3D volumetric mesh, generated by the finite element method
in \cite{petit2015real}. A modified expectation-maximization (EM)
algorithm is designed to directly register a point cloud from an RGB-D
camera to a predefined mechanical model by a physical simulator
in \cite{schulman2013tracking}. However, the physical simulation models only work for linked rigid objects or elastic objects. The accuracy of the algorithms depends
on the physics-based prior and the physical simulation. A physically
accurate model is used in \cite{javdani2011modeling} to model the
elongated deformable object, and the physics-based priors are estimated
by minimizing a generic energy function based on the images from a
calibrated 3-camera rig.

Elongated deformable objects are also modeled by graphs or splines in \cite{DeGregorio, lui2013tangled}. A linear graph (line segments and not the
shape) is used to represent a rope, and particle filters based on
a predefined score function are used to infer the rope configuration
in \cite{lui2013tangled}. A region adjacency graph based on the super-pixels
from the image of wires is developed to model the elongated deformable
object in \cite{DeGregorio}. A method based on topological model
and knot theory is developed to recognize the rope conditions in \cite{matsuno2006manipulation}.
Based on the point cloud from an RGB-D camera, B\'ezier curve chained
rectangles are used to approximate an elongated deformable object,
the corresponding likelihood function is proposed, and the progressive
Gaussian filter is used for the state estimation in \cite{zea2016tracking}.
However, the paper doesn't consider the situation where the elongated
deformable object intersects with itself. A nonuniform rational B-spline
(NURBS) curve is used to model a thin, deformable surgical suture
thread by minimizing the image matching energy between the projected
stereo NURBS image and the segmented thread image \cite{jackson2017real}.

In this paper, an elongated deformable object is modeled as chained
ellipses as shown in Fig. \ref{fig:Rope as MM} (a), where the centers
of the ellipses are located on a B-spline curve (brown dash-dotted
line). Each ellipse (in yellow) is represented by a random matrix
model (RMM), of which the center represents the location and the covariance
matrix represents the shape of the ellipse. The RMM approximation of an elliptical
object is widely used for extended object tracking.
Extended object tracking methods track the position of the centroid and estimate the
shape of the object, given sparse point measurements on the object at each time
frame (cf. \cite{yao2017image,yao2018image,feldmann2011tracking}). The
proposed method is based on the point cloud of the elongated deformable
object obtained by an RGB-D camera. The point cloud generated from
the elongated deformable object is sparse but provides the position
information which can be used as the measurements. 

The B-spline chained RMMs can also be used when both the shape
and the length of the object are changing, e.g. during the manipulations, or when assembling two elongated deformable objects. In addition, the physical
simulation model of the elongated deformable object is not required
to be built beforehand. Technical contributions of the paper
are briefly summarized as follows:
\begin{itemize}
\item A set of chained multiple RMMs is proposed to approximate the elongated
deformable object, of which the centers are enforced to be located
on a B-spline curve. The corresponding likelihood function is derived.
The proposed model both localizes the elongated deformable object
and estimates its shape.
\item A modified EM algorithm is proposed for the shape estimation of the
elongated deformable object, based on the proposed log-likelihood
function. The control points of the B-spline curve, the number of measurement points associated with each RMM and the covariance
matrices of the RMMs are the parameters to be estimated by the EM
algorithm. 
\item Because the log-likelihood function is nonconvex, it is necessary
to have a good initialization for the EM algorithm. Knot sequence
of the B-spline curve also needs to be generated from the unordered
measurement points. A split and merge algorithm is proposed for the
initialization which uses the Euclidean minimum spanning tree (EMST)
and the breadth first search (BFS) method. 
\end{itemize}
The rest of the paper is organized as follows. In Section \ref{sec:Chained-Mixture-Model},
the B-spline curve chained RMMs, its likelihood function and a corresponding EM method are presented to estimate the position and shape of the elongated
deformable object. In Section \ref{sec:Initialization}, a split and
merge method based on the EMST and BFS is proposed to initialize the
chained RMMs. The shape estimation results of the rope in different configurations,
including non-intersection and intersection, based on measurements
from the RGB-D camera, are shown in Section \ref{sec:Experimental-Results}.
The estimation results of the rope as well as the plastic tube in
scenarios such as continuous manipulations and the assembly of two
plastic tubes are also shown in this section. Conclusions and future
work are given in Section \ref{sec:Conclusion}. 
\section{B-spline Chained Random Matrices Model\label{sec:Chained-Mixture-Model}}

\subsection{B-spline Curve Representation}

A point $\mathbf{p}(t)\in\mathbb{R}^{2\times1}$ on the B-spline curve
of degree $d$ can be interpolated with parameter $t$ from a polynomial,
which is defined as a linear combination of $n+1$ control points
(de Boor points) $\mathbf{b}_{i}\in\mathbb{R}^{2\times1}$ and basis
functions $N_{i,d}$ given by \cite{de1978practical}
\begin{equation}
\begin{array}{c}
\begin{aligned}\mathbf{p}(t) & =\sum_{i=0}^{n}\text{\ensuremath{\mathbf{b}}}_{i}N_{i,d}(t)\end{aligned}
\\
0\leq t\leq n-d+1
\end{array}\label{eq:B-spline}
\end{equation}
The basis function $N_{i,d}(t)$ is defined based on a non-decreasing
knot sequence $\left\{ t_{i}\colon i=0\cdots n+d+1\right\} $ as \cite{de1978practical}
\begin{equation}
\begin{aligned}N_{i,d}(t)=\frac{t-t_{i}}{t_{i+d}-t_{i}}N_{i,d-1}(t)+\frac{t_{i+d+1}-t}{t_{i+d+1}-t_{i+1}}N_{i+1,d-1}(t)\end{aligned}
\end{equation}
which is a recursion function and 
\begin{equation}
N_{i,0}(t)=\begin{cases}
1 & t\in\left[t_{i},t_{i+1}\right)\\
0 & otherwise
\end{cases}
\end{equation}
and the knot values $t_{i}$ of the knot sequence are generated by
\begin{equation}
t_{i}=\begin{cases}
0 & 0\leq i<d+1\\
i-d & d+1\leq i\leq n\\
n-d+1 & n<i\leq n+d+1
\end{cases}
\end{equation}

\subsection{Chained Multiple Random Matrices Model}

\begin{figure*}[t]
\begin{centering}
\makebox[0.8\linewidth][c]{\subfigure[]{\includegraphics[height=35mm]{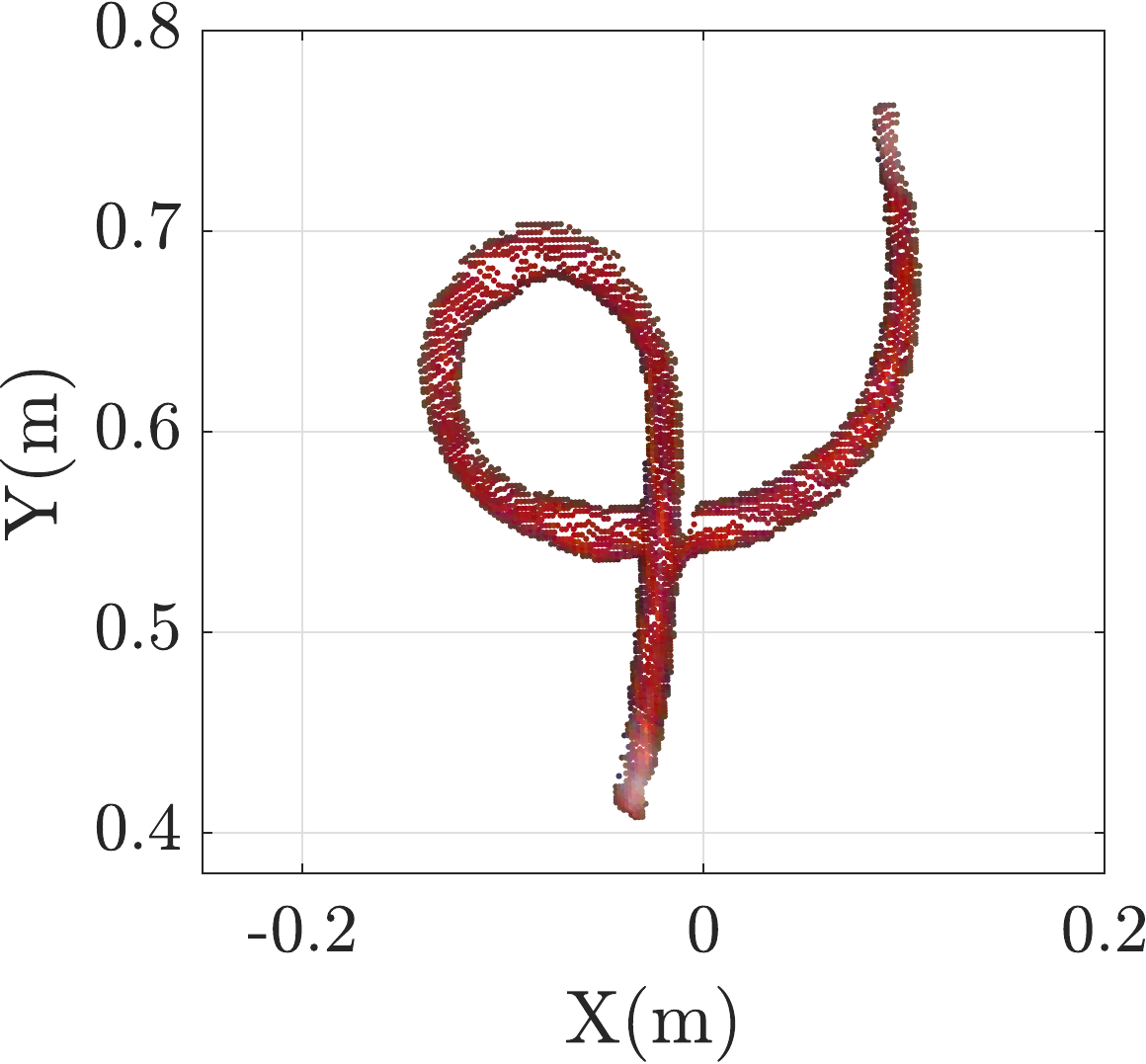}}\hspace{0.5em}\subfigure[]{\includegraphics[height=37mm]{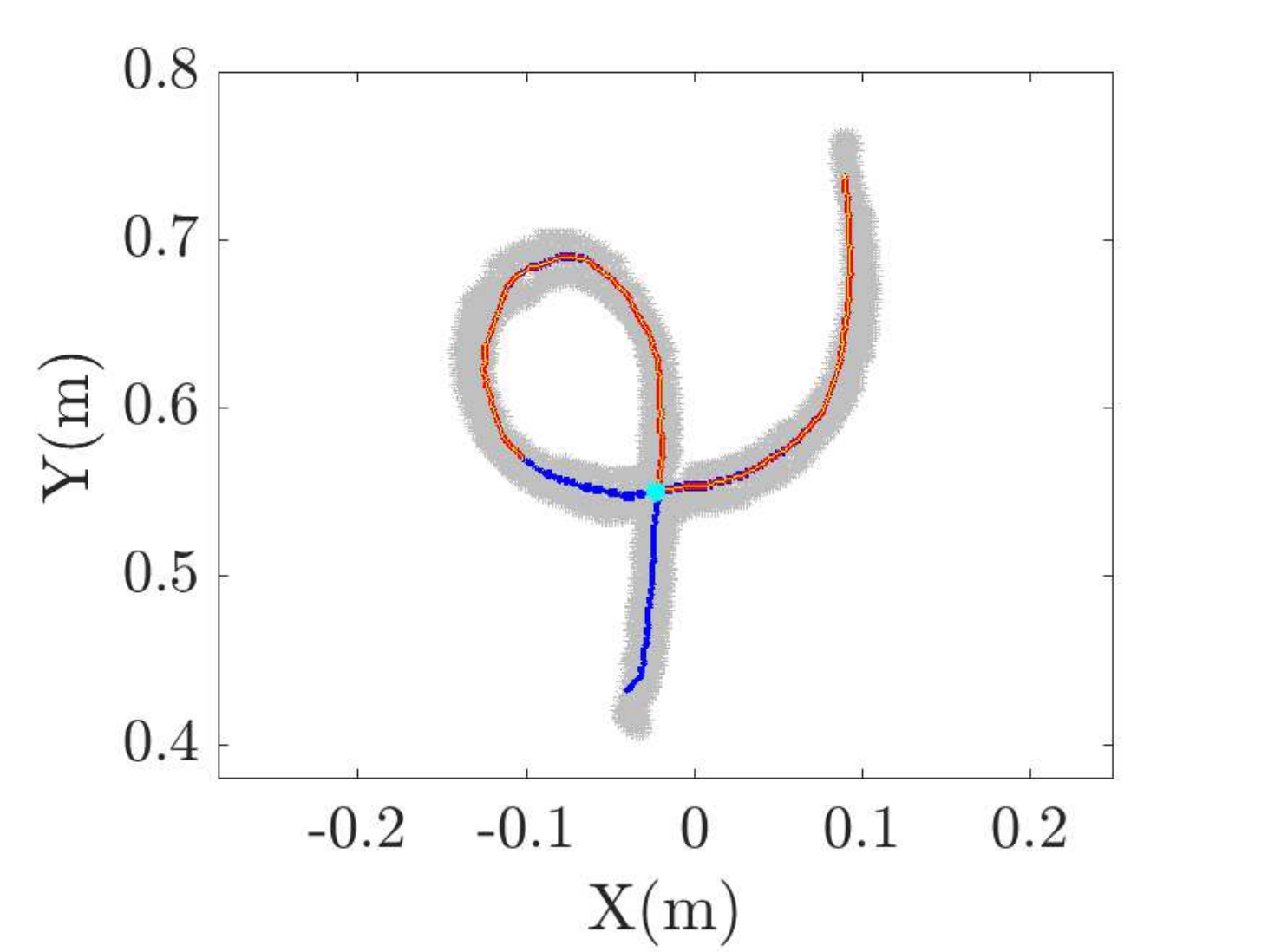}}\hspace{-1em}\subfigure[]{\includegraphics[height=37mm]{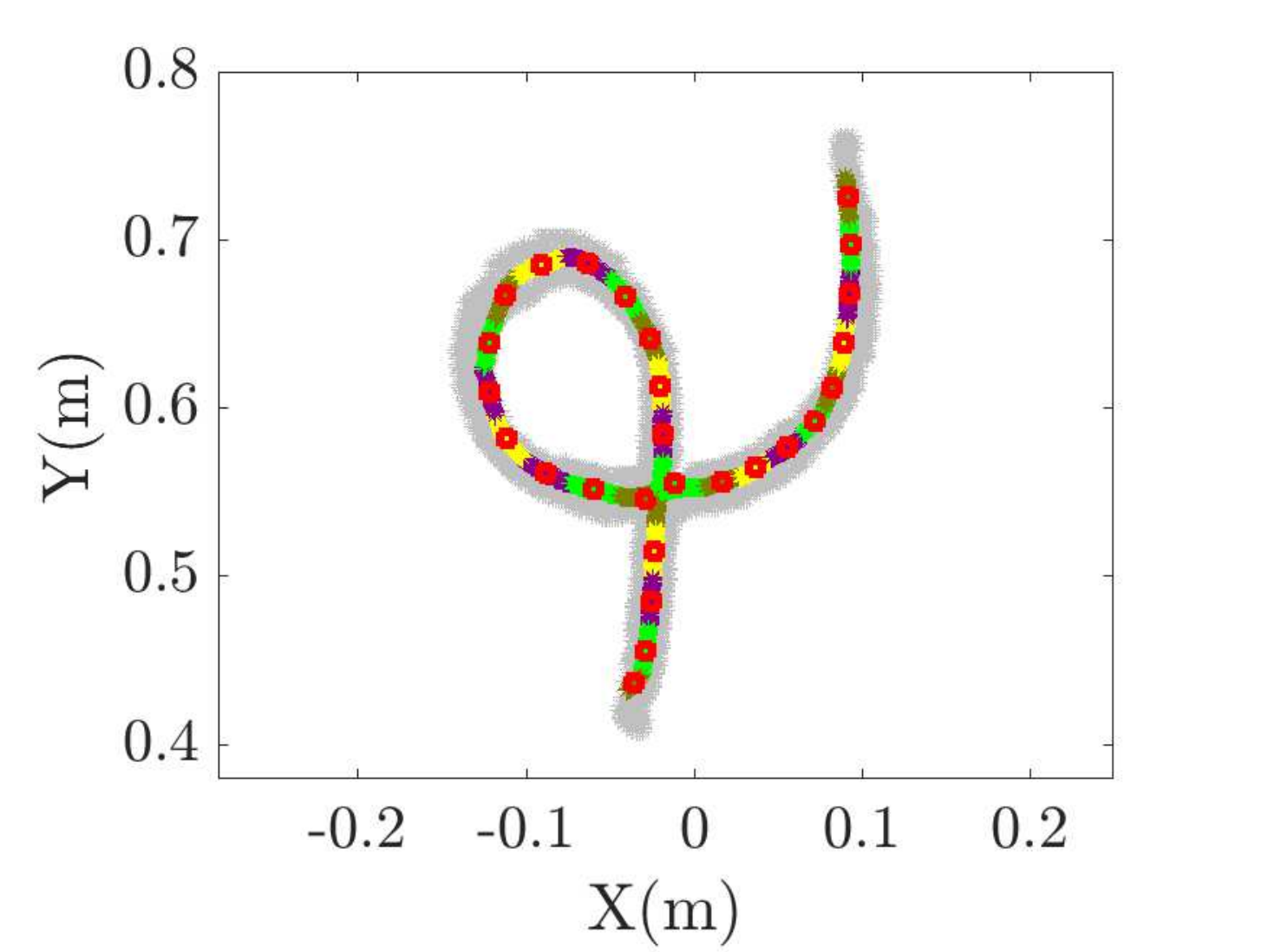}}\hspace{-1em}\subfigure[]{\includegraphics[height=37mm]{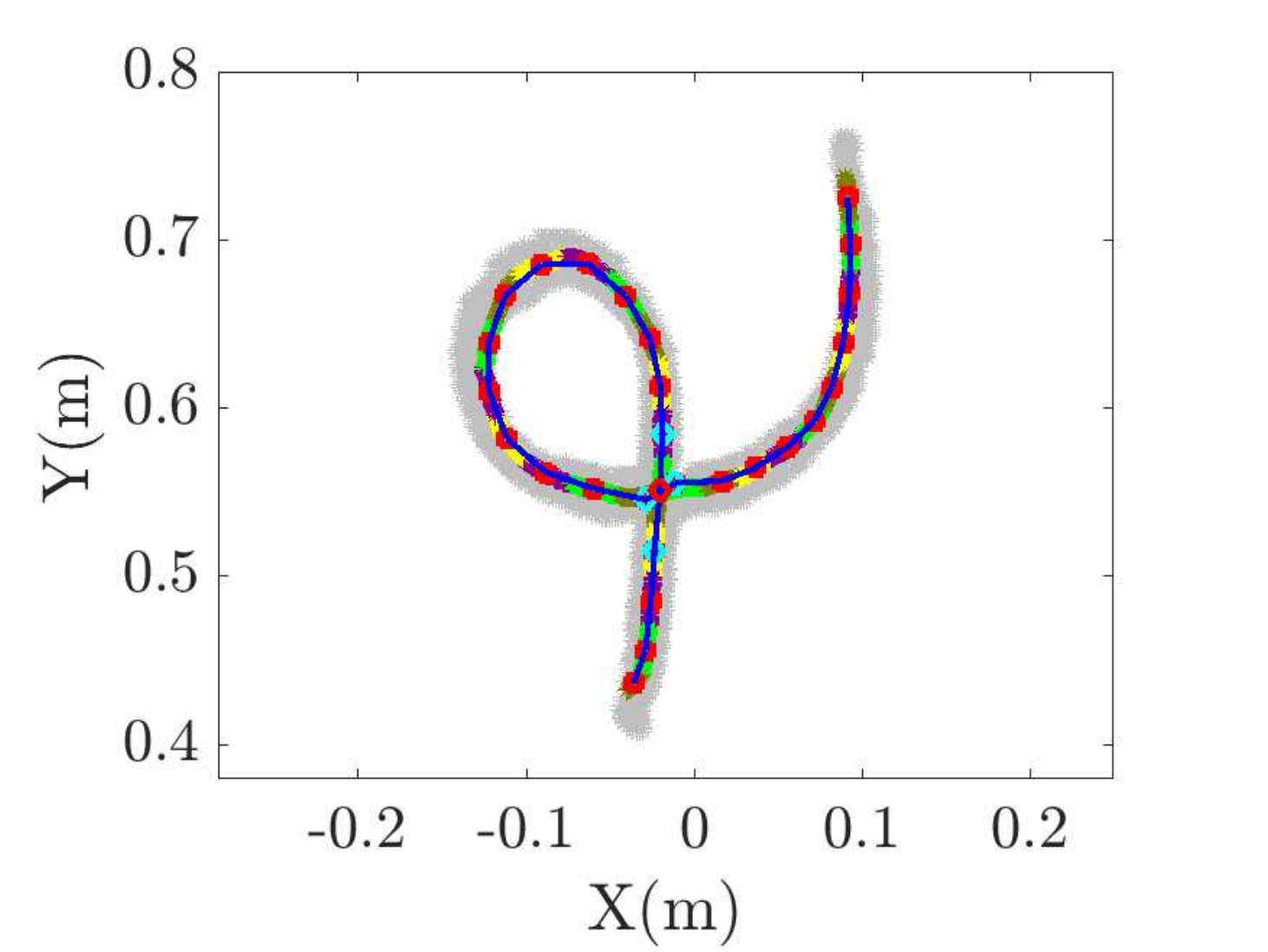}}}\\
\par\end{centering}
\centering{}\caption{Illustration of the initialization procedure: (a) Point cloud from
the rope after background subtraction; (b) EMST of the points,  the
longest path (orange line segments) and the common vertex (in cyan);
(c) Small segments represented by different colors and the corresponding
centers (red points); (d) Graph of centers (blue line segments) and
the common vertex and its 4 closest centers (in cyan). \label{fig:initialization}}
\end{figure*}

The B-spline chained multiple random matrices model is used to model
the elongated deformable object in this section as shown in Fig. \ref{fig:Rope as MM}.
The multiple random matrices model is the sum of equally weighted
$K$ RMMs, which is defined as
\begin{equation}
p(\mathbf{Z}\mid\theta)=\sum_{k=1}^{K}w_{k}\phi(\mathbf{Z}_{k}|L_{k},\mu_{k},\Sigma_{k})\label{eq:RMMs}
\end{equation}
 where $\theta=\left\{ \left(L_{k},\mu_{k},\Sigma_{k}\right)\right\} _{k=1}^{K}$,
$L_{k}$ is the number of measurements assigned to the $k\textrm{th}$
RMM, $\mu_{k}$ and $\Sigma_{k}$ are the mean and covariance of the
$k\textrm{th}$ cluster, $\phi(\mathbf{Z}_{k}|L_{k},\mu_{k},\Sigma_{k})$
is the probability distribution of the measurement points $\mathbf{Z}_{k}=\{\mathbf{z}_{k,l}\}_{l=1}^{L_{k}}$
from the $k\textrm{th}$ cluster, $\mathbf{Z}=\{\mathbf{z}_{r}\}_{r=1}^{N_{r}}$
is the set of total number of measurement points, and the weights
of the clusters are assumed to be equal as $w_{k}=\frac{1}{K}$. 

The $k\textrm{th}$ cluster is approximated as an ellipse, and the
measurement points inside the ellipse are assumed to be distributed
as a Gaussian distribution with mean $\mu_{k}$ and covariance $\Sigma_{k}$.
The probability of the measurement points $\mathbf{Z}_{k}=\{\mathbf{z}_{k,l}\}_{l=1}^{L_{k}}$
from the $k\textrm{th}$ ellipse represented by RMM is \cite{feldmann2011tracking} 

\begin{equation}
\begin{aligned}\phi(\mathbf{Z}_{k}|L_{k},\mu_{k},\Sigma_{k}) & =\prod_{l=1}^{L_{k}}\mathcal{N}(\mathbf{z}_{k,l};\mu_{k},\Sigma_{k})\\
 & \propto\mathcal{N}(\mathbf{\overline{z}}_{k};\mu_{k},\frac{\Sigma_{k}}{L_{k}})\times W(\mathbf{\overline{C}}_{k};L_{k}-1,\Sigma_{k})
\end{aligned}
\label{eq:RMM}
\end{equation}
where the center $\mathbf{\overline{z}}_{k}$ is

\begin{equation}
\mathbf{\overline{z}}_{k}=\frac{1}{L_{k}}\sum_{l=1}^{L_{k}}\mathbf{z}_{k,l}\label{eq:center}
\end{equation}
and the scattering matrix $\mathbf{\overline{C}}_{k}$ is

\begin{equation}
\mathbf{\overline{C}}_{k}=\sum_{l=1}^{L_{k}}(\mathbf{z}_{k,l}-\mathbf{\overline{z}}_{k,l})(\mathbf{z}_{k,l}-\mathbf{\overline{z}}_{k,l})^{\mathrm{T}}\label{scattering matrix}
\end{equation}
and $W(\mathbf{\overline{C}}_{k};L_{k}-1,\Sigma_{k})$ is a Wishart
density in $\mathbf{\overline{C}}_{k}$ with $L_{k}-1$ degrees of
freedom. The statistical sensor errors are assumed to be neglected,
and the physical extension of the target dominates the spread of measurements
in (\ref{eq:RMM}) \cite{feldmann2011tracking}. If the sensor errors
are within the same order of the magnitude of the target extension,
they cannot be neglected anymore. In this case, the covariance is
$\Sigma_{k}=\Sigma_{p}+\boldsymbol{\mathbf{R}}$, where $\Sigma_{p}$
is a symmetric positive definite random matrix representing the physical
extension, and $\mathbf{R}$ is the covariance matrix of sensor errors
\cite{feldmann2011tracking}.

The brown dash-dotted curve in Fig. \ref{fig:Rope as MM} (a) is the
B-spline curve, which crosses through the centers of the clusters.
Multiple RMMs with centers located on the B-spline curve constitute
the elongated deformable object. The probability of the measurement
points $\mathbf{Z}_{k}=\{\mathbf{z}_{k,l}\}_{l=1}^{L_{k}}$ from the
$k\textrm{th}$ elliptical cluster is defined as
\begin{equation}
\begin{aligned}\phi(\mathbf{Z}_{k}|L_{k},\text{\ensuremath{\mathbf{c}}}_{k},\Sigma_{k})= & \intop_{\mu_{k}}\phi(\mathbf{Z}_{k}|L_{k},\mu_{k},\Sigma_{k})\cdot p(\mu_{k}|\mathbf{c}_{k})d\mu_{k}\end{aligned}
\label{eq:MRMM}
\end{equation}
where $\phi(\mathbf{Z}_{k}|L_{k},\text{\ensuremath{\mathbf{c}}}_{k},\Sigma_{k})$
is the probability distribution of the measurement points from the
$k\textrm{th}$ cluster and $p(\mu_{k}|\mathbf{c}_{k})=\delta(\mu_{k}-\mathbf{c}_{k})$
is the Dirac delta function to enforce the center $\mu_{k}$ to be
located on the B-spline curve $\mathbf{c}_{k}$. $K$ center points
are sampled from the B-spline curve as
\begin{equation}
\begin{aligned}\mathbf{c}_{k}=\mathbf{p}(t_{k}),\; & k=1\cdots K\end{aligned}
\label{eq:Bsplinecenter}
\end{equation}
where $\mathbf{c}_{k}$ is the $k\textrm{th}$ center point, and $\mathbf{p}(t_{k})$
is the corresponding point on the B-spline curve, which is only determined
by the control points of the B-spline curve $\text{\ensuremath{\mathbf{b}}}_{i}$
in (\ref{eq:B-spline}), and $t_{k}$ is the corresponding parameter
value of the B-spline curve determined by the centripetal method \cite{lee1989choosing}. 

Assuming the $k\textrm{th}$ cluster center $\mathbf{c}_{k}$ is located
on the B-spline curve, by putting (\ref{eq:RMM}) and (\ref{eq:Bsplinecenter})
into (\ref{eq:MRMM}) the probability distribution of the measurement
points from the $k\textrm{th}$ cluster is 
\begin{equation}
\phi(\mathbf{Z}_{k}|L_{k},\text{\ensuremath{\mathbf{c}}}_{k},\Sigma_{k})\propto\mathcal{N}(\mathbf{\overline{z}}_{k,l};\mathbf{c}_{k},\frac{\Sigma_{k}}{L_{k}})\times W(\mathbf{\overline{C}}_{k};L_{k}-1,\Sigma_{k})
\end{equation}
The B-spline curve chained multiple random matrices representation is the sum of equally weighted
$K$ RMMs, which is redefined as
\begin{equation}
p(\mathbf{Z}\mid\theta)=\sum_{k=1}^{K}w_{k}\phi(\mathbf{Z}_{k}|L_{k},\text{\ensuremath{\mathbf{c}}}_{k},\Sigma_{k})\label{eq:RMMs-1}
\end{equation}

\subsection{Expectation-Maximization Method}

In this subsection, the EM algorithm is used to estimate the parameters
$\theta=\left\{ \left(L_{k},\text{\ensuremath{\mathbf{c}}}_{k},\Sigma_{k}\right)\right\} _{k=1}^{K}$
of the B-spline chained RMMs. The EM algorithm finds the parameters
$\theta^{*}=\left\{ \left(L_{k}^{*},\text{\ensuremath{\mathbf{c}}}_{k}^{*},\Sigma_{k}^{*}\right)\right\} _{k=1}^{K}$
corresponding to the maximum likelihood by iterating between the expectation
step and the maximization step. 

The \textbf{expectation step} assigns each of the $N_{r}$ measurement
points $\mathbf{Z}=\{\mathbf{z}_{r}\}_{r=1}^{N_{r}}$ to each of clusters
by 
\begin{equation}
k^{*}=\underset{k=1\cdots K}{\mathrm{arg\,max}}\,p_{k}(\mathbf{z}_{r})\label{eq:expectation step}
\end{equation}
where $p_{k}(\mathbf{z}_{r})=\mathcal{N}(\mathbf{z}_{r};\text{\ensuremath{\mathbf{c}}}_{k},\Sigma_{k})$.
The parameter $L_{k}$ is then determined by counting the number of
points in each cluster. After the assignment of the measurements,
the mean $\mathbf{\overline{z}}_{k}$ and scattering matrix $\mathbf{\overline{C}}_{k}$
are calculated using (\ref{eq:center}) and (\ref{scattering matrix}).
The corresponding parameter $t_{k}$ in (\ref{eq:Bsplinecenter})
is also recalculated based on the centripetal method \cite{lee1989choosing}. 

The \textbf{maximization step} estimates the parameters $\theta$
by maximizing the log-likelihood function of the chained RMMs. The
log-likelihood function of the chained RMMs in (\ref{eq:RMMs-1})
is
\begin{equation}
\begin{aligned}\mathscr{\mathcal{L}}(\theta)= & \sum_{k=1}^{K}\left\{ -\frac{L_{k}}{2}(\text{\ensuremath{\mathbf{c}}}_{k}-\mathbf{\overline{z}}_{k})^{\mathrm{T}}\Sigma_{k}^{-1}(\text{\ensuremath{\mathbf{c}}}_{k}-\mathbf{\overline{z}}_{k})\right.\\
 & \left.-\frac{L_{k}+1}{2}\log\left|\Sigma_{k}\right|-\frac{1}{2}\textrm{tr}(-\frac{1}{2}\mathbf{\overline{C}}_{k}\Sigma_{k}^{-1})\right\} +\textrm{Const}
\end{aligned}
\label{eq:negativelotlike}
\end{equation}
which is maximized by iterative re-weighted least squares method \cite{bishop2006pattern}.
Rewrite (\ref{eq:B-spline}) in the matrix-vector form as $\text{\ensuremath{\mathbf{c}}}_{k}=\mathbf{B}_{k}\mathbf{b}$,
where $\mathbf{B}_{k}$ is the block diagonal matrix as $\mathbf{B}_{k}=\mathbf{blkdiag}(\mathbf{n}_{k}^{\mathrm{T}},\mathbf{n}_{k}^{\mathrm{T}})\in\mathbb{R}^{2\times2(n+1)}$,
$\mathbf{n}_{k}=\left[\begin{array}{ccc}
N_{0,d}(t_{k}), & \cdots & N_{n,d}(t_{k})\end{array}\right]^{\mathrm{T}}\in\mathbb{R}^{(n+1)\times1}$, and $\mathbf{b}=\left[\begin{array}{cc}
\mathbf{b}_{x}^{\mathrm{T}}, & \mathbf{b}_{y}^{\mathrm{T}}\end{array}\right]^{\mathrm{T}}\in\mathbb{R}^{2(n+1)\times1}$, where $\mathbf{b}_{x}^{\mathrm{T}}$ and $\mathbf{b}_{y}^{\mathrm{T}}$
are the control points in x and y coordinates. Taking the derivative
of the log-likelihood function $\mathcal{L}(\theta)$ with respect
to the control points $\text{\ensuremath{\mathbf{b}}}$ and positive
symmetric matrix $\Sigma_{k}$ separately and setting them equal to
0 yields
\begin{equation}
\mathbf{b}=\mathbf{Q}^{+}\mathbf{M}\label{eq:optimal-center}
\end{equation}
where $\mathbf{Q^{+}}$ is the Moore--Penrose inverse of $\mathbf{Q}=\sum_{k=1}^{K}L_{k}\mathbf{B}_{k}^{\mathrm{T}}\Sigma_{k}^{-1}\mathbf{B}_{k}$
and $\mathbf{M}=\sum_{k=1}^{K}L_{k}\mathbf{B}_{k}^{\mathrm{T}}\Sigma_{k}^{-1}\mathbf{\overline{z}}_{k}$,
and 
\begin{equation}
\Sigma_{k}=\frac{1}{L_{k}+1}\sum_{l=1}^{L_{k}}(\mathbf{z}_{k,l}-\mathbf{c}_{k})(\mathbf{z}_{k,l}-\mathbf{c}_{k})^{\mathrm{T}}\label{eq:optimal-sigma}
\end{equation}

The iteration between (\ref{eq:optimal-center}) and (\ref{eq:optimal-sigma})
is carried out until the value of the log-likelihood function in (\ref{eq:negativelotlike})
stops increasing or the optimization reaches the predefined maximum
iteration number. 

The orientation of the ellipse and its semi-major (red arrow) and
semi-minor (green arrow) axes are determined by the eigenvalues and
eigenvectors of $4\Sigma_{k}$ as shown in Fig. \ref{fig:Rope as MM}
(a), assuming that the measurements are uniformly distributed inside
the ellipse and approximated as a Gaussian distribution in the proposed
model by the moment matching method \cite{feldmann2011tracking}. 

In every ellipse, each line perpendicular to the B-spline curve that
passes through the center is found. The length of the line segment
(blue line segment as shown in Fig. \ref{fig:Rope as MM} (a)) between
the center and its intersection with the ellipse is calculated. Then,
half of the width of the rope is determined by the average length
of the calculated line segments. The offset curves are drawn by off-shifting
the B-spline curve by half of the calculated width of the rope. 

The two ends ($\mathbf{c_{1}}$ and $\mathbf{c}_{K}$ in Fig. \ref{fig:Rope as MM}
(a)) of the B-spline curve are the centers of the ellipses representing
the two terminal parts of the rope. The two ends ($\mathbf{e_{1}}$
and $\mathbf{e}_{2}$ in Fig. \ref{fig:Rope as MM} (a)) of the rope
are determined by the intersections between these ellipses and the
lines (red dashed lines in Fig. \ref{fig:Rope as MM} (a)) tangent
to the B-spline curve at the centers ($\mathbf{c_{1}}$ and $\mathbf{c}_{K}$
in Fig. \ref{fig:Rope as MM} (a)) of these corresponding ellipses.
The length of the rope is determined by the length of the B-spline
curve and the lengths of the line segments (red dashed lines in Fig.
\ref{fig:Rope as MM} (a)) between the two ends of the rope and the
centers of the corresponding ellipses.

\section{Initialization for B-spline Chained RMMs\label{sec:Initialization}}

The elongated deformable object may have parts which are very close
to one another or have intersections with itself. The initialization
step is to generate the general configuration of the elongated deformable
object from the point cloud and to trace a B-spline curve embedded
in the unordered measurement points. A split and merge method is proposed
to initialize the algorithm in this section. The rope is first split
into small segments and then the configuration of the rope is obtained
by building the graph of the centers of the small segments. The initialization
procedure is shown in Fig. \ref{fig:initialization}.

\begin{figure}
\begin{centering}
\makebox[0.8\linewidth][c]{\subfigure[]{\includegraphics[height=35mm]{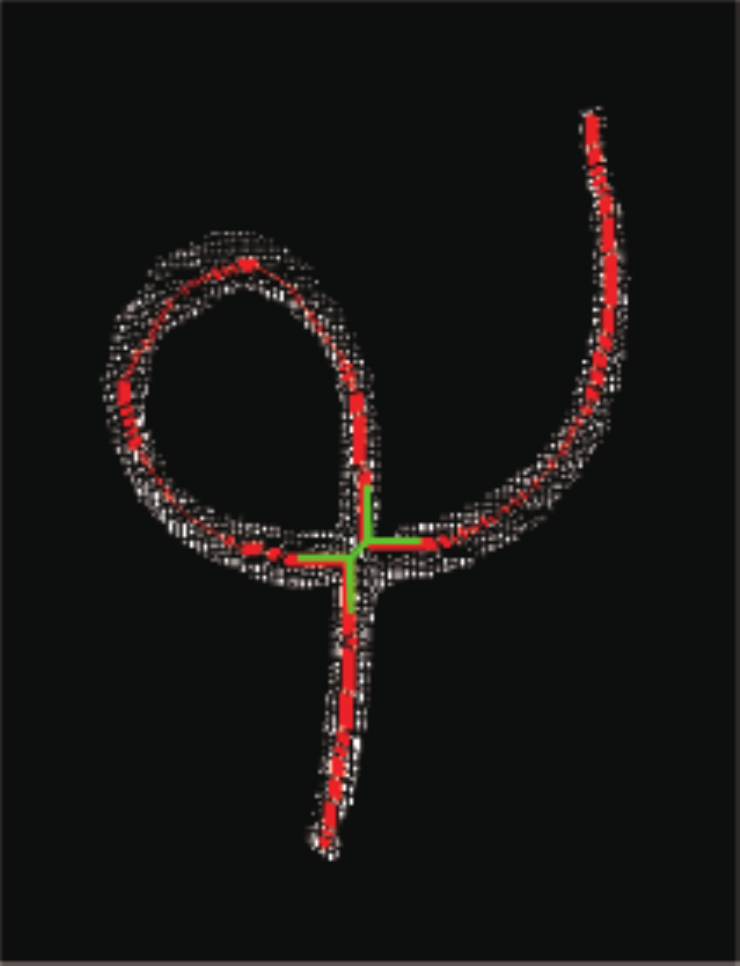}}\hspace{2em}\subfigure[]{\includegraphics[height=35mm]{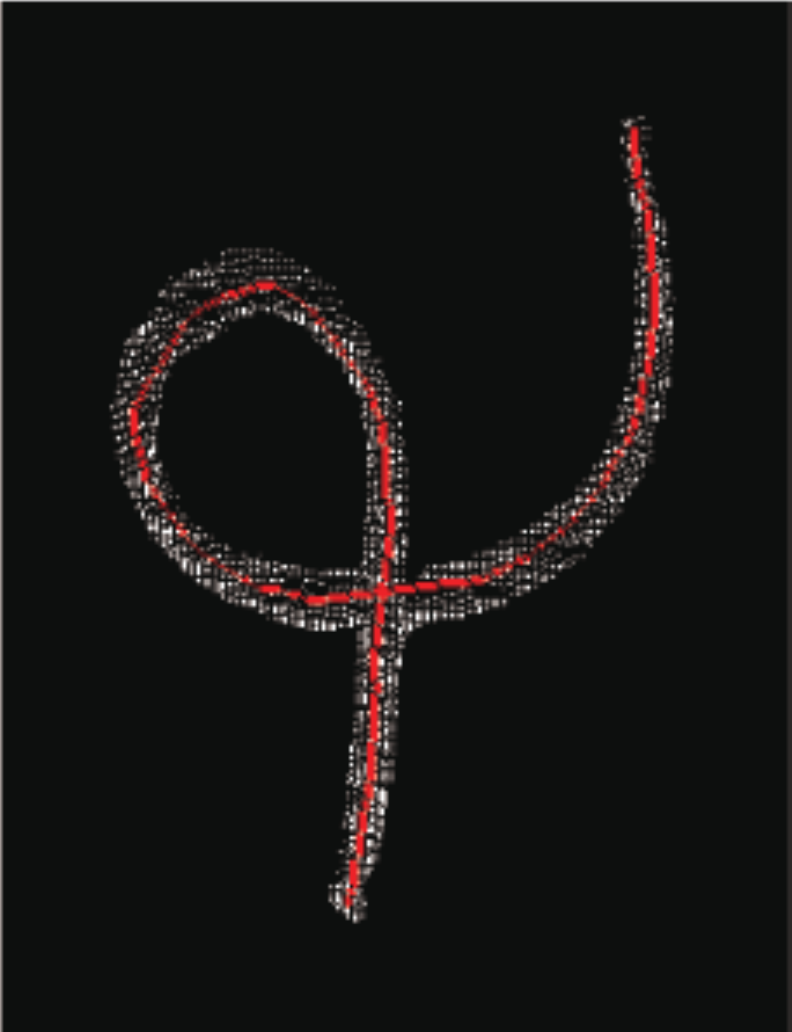}}}\\\caption{The binary image of the point cloud (in white) from the rope and the
linearization of the intersection points: (a) The pixels (in red)
obtained by the dilation and thinning of the binary image, and the
intersection points (in green); (b) The pixels linearized by the Bresenham
algorithm. \label{fig:The-binary-image}}
\par\end{centering}
\end{figure}

Before the initialization procedure, the pre-processing stage is done
to find the medial skeleton of the point cloud. The point cloud of
the rope after background subtraction is shown in Fig. \ref{fig:initialization}
(a). Then, the point cloud is converted into a binary image which
is then dilated and thinned. The pixels (in red) after dilation and
thinning are obtained as shown in Fig. \ref{fig:The-binary-image}.
The intersection part of the rope (in green) is linearized by the
Bresenham algorithm (see \cite{matsuno2006manipulation}). Then the
pixels are converted back to the point cloud with position information
as shown in Fig. \ref{fig:initialization} (b). Other methods to find
the medial skeleton of the point cloud can also be used in this stage
\cite{huang2013l1}. 

\subsection{Split Step}

The first step of the split and merge method is to divide the obtained
point cloud into smaller segments, as shown in Algorithm \ref{alg:The-segmentation-algorithm}.
First, an EMST is constructed based on the points. The EMST is an
acyclic edge-weighted graph $T=(\mathcal{V},\mathcal{E})$, where
$\mathcal{V}$ is the vertices set and $\mathcal{E}$ is the set of
the edges connecting every two vertices $\mathbf{v}_{i},\mathbf{v}_{j}\in\mathcal{V}$
and $i\neq j$. The weight of the edge is defined as the Euclidean
distance of the two vertices $\epsilon=\left\Vert \mathbf{v}_{i}-\mathbf{v}_{j}\right\Vert $
\cite{lee2000curve,sedgewick2011algorithms}. The EMST is constructed
by selecting the set of connected edges to ensure the summation of
the weights of the edges is minimum. The Prim's algorithm is used
to generate the EMST \cite{sedgewick2011algorithms}. 

After the construction of the EMST, the longest path of the EMST is
found by two breadth first searches (BFS) \cite{sedgewick2011algorithms}.
The first BFS is used to traverse the EMST with a random chosen vertex
from the EMST, and the path with the largest weight and one end point
of the longest path are found. Then, another BFS traverses the EMST,
starting with the end point that was found in the prior iteration.
The longest path $P=(\mathcal{V}_{p},\mathcal{E}_{p})$ is found which
is also the longest path of the EMST $T=(\mathcal{V},\mathcal{E})$,
where $\mathcal{V}_{p}\subset\mathcal{V}$ and $\mathcal{E}_{p}\subset\mathcal{E}$.
The orange line segments shown in Fig. \ref{fig:initialization} (b)
are the longest path of the EMST $T=(\mathcal{V},\mathcal{E})$. 

\begin{figure}[H]
\centering{}\includegraphics[width=0.7\columnwidth]{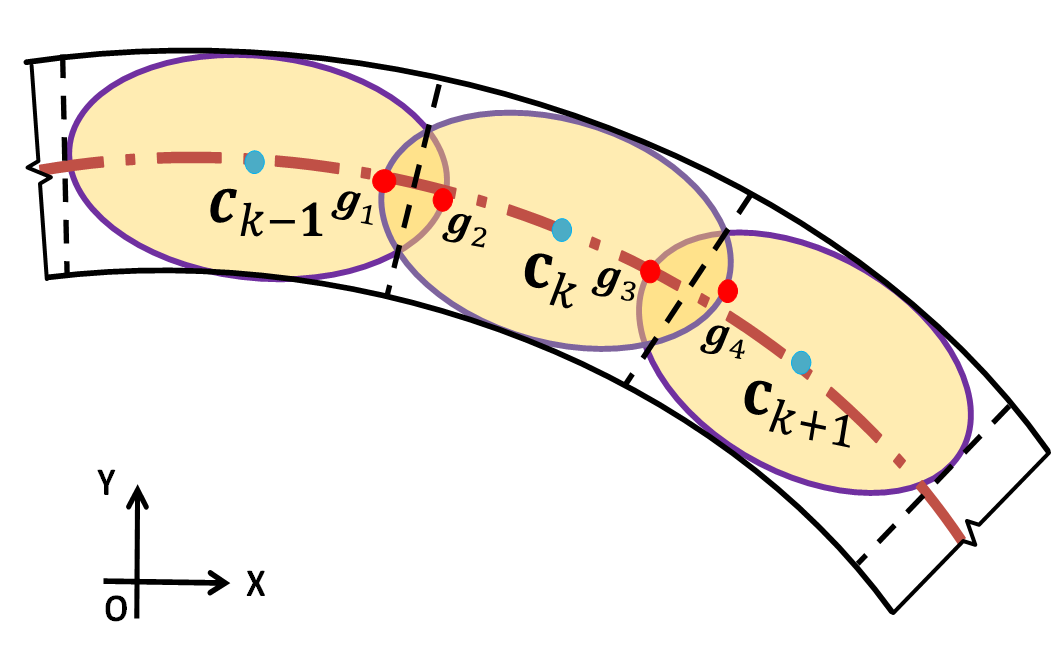}\caption{Adjustment of the ellipses with overlaps.\label{fig:overlaps}}
\end{figure}

\begin{algorithm}
Set the threshold $\gamma$\; 

Obtain the point cloud $\mathbf{Z}=\{\mathbf{z}_{r}\}_{r=1}^{N_{r}}$\; 

Build the EMST $T=(\mathcal{V},\mathcal{E})$\; 

Find the longest path of the EMST $P=(\mathcal{V}_{p},\mathcal{E}_{p})$
by BFSs\; 

Delete $\mathcal{E}_{p}$, then there are many small clusters (trees)
$\mathcal{C}=\left\{ \mathcal{C}_{h}\right\} _{h=1}^{J_{h}}$\; 

Initialize empty sets $\mathcal{A}$ and $\mathcal{B}$\; 

Initialize $h=1$\; 

\While{ $h\leq J_{h}$ }{

Calculate $D_{h}$ in (\ref{eq:dsitance})\; 

\If{ $D_{h}>\gamma$}{ 

Append $\mathcal{C}_{h}$ to $\mathcal{A}$\;}

\Else{$\mathcal{B}=\mathcal{B}\cup\mathcal{C}_{h}$\;}

$h=h+1$\;}

Append $\mathcal{B}$ to $\mathcal{A}$\; 

\caption{The segmentation of the elongated deformable object\label{alg:The-segmentation-algorithm} }
\end{algorithm}

After the construction of the EMST and the longest path $P=(\mathcal{V}_{p},\mathcal{E}_{p})$
is found, the EMST is segmented into smaller disconnected clusters
(or trees) by deleting the $\mathcal{E}_{p}$ from the EMST $T$.
Then, there are some small trees $\mathcal{C}=\left\{ \mathcal{C}_{h}\right\} _{h=1}^{J_{h}}$
left. The distance $D_{h}$ between $\mathcal{C}_{h}$ and $\mathcal{V}_{p}$
is defined as
\begin{equation}
D_{h}=\textrm{sup}\left\{ \textrm{inf}\{\left\Vert \mathbf{u}_{i}-\mathbf{v}_{j}\right\Vert |\forall\mathbf{v}_{j}\in\mathcal{V}_{p}\}|\forall\mathbf{u}_{i}\in\mathcal{C}_{h}\right\} \label{eq:dsitance}
\end{equation}
where $\mathbf{v}_{j}$ are vertices from $\mathcal{V}_{p}$ and $\mathbf{u}_{i}$
are vertices from $\mathcal{C}_{h}$. If $D_{h}$ is larger than a
predefined threshold $\gamma$, the cluster $\mathcal{C}_{h}$ and
the $\mathcal{V}_{p}$ are from different regions of the rope, instead
of the small branches of the $P$. After the $O_{n}$ different parts
(or smaller clusters) of the rope $\mathcal{A}=\{\mathcal{A}_{n}\}_{n=1}^{O_{n}}$
are found, the common vertices $\mathcal{V}_{C}=\{\mathbf{v}_{c}|\mathbf{v}_{c}\in\mathcal{A}\backslash\mathcal{B},\mathbf{v}_{c}\in\mathcal{V}_{P},c=1,\cdots,O_{n}-1\}$
of the $O_{n}$ clusters are also found, where the points of $\mathcal{B}$
are from the same region of the rope as $\mathcal{V}_{P}$. The common
vertices have more than $2$ edges connected with them. If the distance between some common vertices are smaller than the predefined threshold $\gamma$, their mean values are used to represent them. The EMST is
segmented into $O_{n}=2$ clusters (orange and blue line segments)
shown in Fig. \ref{fig:initialization} (b). 

Each cluster is an EMST and the longest path with two end points are
found by two BFSs. Starting with one end point, the cluster is divided
into $M_{n}$ smaller segments $\mathcal{S}_{n}=\{\mathcal{S}_{u,n}\}_{u=1}^{M_{n}}\subset\mathcal{A}_{n}$,
with each segment $\mathcal{S}_{u,n}$ satisfying
\begin{equation}
\textrm{sup}\{\left\Vert \mathbf{z}_{i}-\mathbf{z}_{j}\right\Vert |\forall\mathbf{z}_{i},\mathbf{z}_{j}\in\mathcal{S}_{u,n},i\neq j\}\leq H\label{eq:segment-criteria}
\end{equation}
where $H$ is a predefined parameter and $\mathbf{z}$ is the point
inside the smaller segments $\mathcal{S}_{u,n}$. The center of each
segment $\mathbf{c}_{u,n}$ is calculated as
\begin{equation}
\mathbf{c}_{u,n}=\frac{1}{L_{u,n}}\sum_{i=1}^{L_{u,n}}\mathbf{z}_{i},\mathbf{z}_{i}\in\mathcal{S}_{u,n}\label{eq: cluster center}
\end{equation}
where $L_{u,n}$ is the number of points in segment $\mathcal{S}_{u,n}$.
The segments are shown with different colors and the red points are
the centers of the segments in Fig. \ref{fig:initialization} (c).
\begin{algorithm}[h]
Set the distance $H$\; 

Given the $O_{n}$ clusters $\mathcal{A}=\{\mathcal{A}_{n}\}_{n=1}^{O_{n}}$\; 

Given the common vertices $\mathcal{V}_{C}=\{\mathbf{v}_{c}|\mathbf{v}_{c}\in\mathcal{\mathcal{A}\backslash\mathcal{B}},\mathbf{v}_{c}\in\mathcal{V}_{p},c=1,\cdots,O_{n}-1\}$\; 

Initialize $n,c=1$\; 

\While{$n\leq O_{n}$ }{

Divide $\mathcal{A}_{n}$ into smaller segments $\mathcal{S}_{n}=\{\mathcal{S}_{u,n}\}_{u=1}^{M_{n}}$
satisfies (\ref{eq:segment-criteria})\; 

Calculate $\mathbf{c}_{u,n}$ in (\ref{eq: cluster center})\; 

Create the graph $T_{m}=(\mathcal{V}_{m},\mathcal{E}_{m})$ with $\mathbf{c}_{u,n}\in\mathcal{V}_{m}$
and $\{\mathbf{c}_{u,n},\mathbf{c}_{u+1,n}\}\in\mathcal{E}_{m}$ and
$u=1,\cdots,M_{n}-1$\; 

$n=n+1$\;}

\While{$c\leq O_{n}-1$ }{

Find the centers $\mathcal{Q}=\{\mathbf{c}_{q}\}_{q=1}^{4}\subset\mathcal{V}_{m}$
closest to $\mathbf{v}_{c}$, and delete edges between them\; 

Calculate vectors $\mathbf{d}_{q}$\; 

Find the two pairs $(i^{*},j^{*})$, based on (\ref{eq:order-criteria})\; 

Create the edges $e$\; 

$\mathcal{E}_{m}=\mathcal{E}_{m}\cup e$\;

$\mathcal{V}_{m}=\mathcal{V}_{m}\cup\mathbf{v}_{c}$\;

$c=c+1$\;}

Complete the graph $T_{m}=(\mathcal{V}_{m},\mathcal{E}_{m})$\;

Fit B-spline by the ordered centers and readjust the centers\; 

\caption{The ordering of the centers and B-spline interpolation\label{alg:merge} }
\end{algorithm}

\subsection{Merge Step}

Previously the elongated target is subdivided into small segments
and the centers of the segments are found. In this step, the order
information of the centers is generated and a B-spline curve is traced,
which represents the global shape and configuration information of
the target as shown in Algorithm \ref{alg:merge}. At first, a graph
$T_{m}=(\mathcal{V}_{m},\mathcal{E}_{m})$ is created, such that the
center of each segment constitutes the vertex $\mathbf{c}_{u,n}\in\mathcal{V}_{m}$
and the edges are built by connecting the centers of nearby segments
in each cluster $\mathcal{A}_{n}$ as $\{\mathbf{c}_{u,n},\mathbf{c}_{u+1,n}\}\in\mathcal{E}_{m}$
and $u=1,\cdots,M_{n}-1$. 

For the common vertex $\mathbf{v}_{c}\in\mathcal{V}_{C}$, the closest
four centers are found $\mathcal{Q}=\{\mathbf{c}_{q}\}_{q=1}^{4}\subset\mathcal{V}_{m}$.
The edges between these four centers are deleted and the new edges
between them are found in the following way. The vectors $\mathbf{d}_{q}=\frac{\mathbf{v}_{c}-\mathbf{c}_{q}}{\left\Vert \mathbf{v}_{c}-\mathbf{c}_{q}\right\Vert }$
are calculated, and the pair $(i^{*},j^{*})$ is found that meets
\begin{equation}
\textrm{\ensuremath{\underset{\mathit{i,j\in q,i\neq j}}{\mathrm{arg\,}\min}}}\left\Vert \mathbf{d}_{i}+\mathbf{d}_{j}\right\Vert \label{eq:order-criteria}
\end{equation}
The vertex $\mathbf{v}_{c}$ is added to the graph as $\mathcal{V}_{m}=\mathcal{V}_{m}\cup\mathbf{v}_{c}$.
Then, the edges are generated as $e_{1}=\{\mathbf{c}_{i^{*}},\mathbf{v}_{c}\}$
and $e_{2}=\{\mathbf{v}_{c},\mathbf{c}_{j^{*}}\}$, which connect
between $\mathbf{v}_{c}$ and the two centers $\mathbf{c}_{i^{*}}$
and $\mathbf{c}_{j^{*}}$, giving the minimum value of (\ref{eq:order-criteria}).
The edges $e_{1}$ and $e_{2}$ are added to the graph $T_{m}$ as
$\mathcal{E}_{m}=\mathcal{E}_{m}\cup e_{1}$ and $\mathcal{E}_{m}=\mathcal{E}_{m}\cup e_{2}$.
The $2$ remaining centers are connected as another edge $e_{3}$
and it is added to the graph $T_{m}$ as $\mathcal{E}_{m}=\mathcal{E}_{m}\cup e_{3}$.
Finally, the graph $T_{m}=(\mathcal{V}_{m},\mathcal{E}_{m})$ is completed
by adding an edge to the $2$ centers of the closest end segments
from different clusters to make all the centers ordered. 

The initial B-spline curve is interpolated by the ordered centers
using (\ref{eq:optimal-center}) and (\ref{eq:optimal-sigma}). The
ellipses may have overlaps as shown in Fig. \ref{fig:overlaps}. The
centers are adjusted to decrease the overlap between ellipses by using
the following formula: 
\begin{equation}
\mathbf{c}_{k}=\frac{\left(\mathbf{g}_{1}+\mathbf{g}_{2}\right)+\left(\mathbf{g}_{3}+\mathbf{g}_{4}\right)}{4}
\end{equation}
where $\mathbf{c}_{k}$ is the center of the $k\textrm{th}$ ellipse,
$\mathbf{g}_{1}$ and $\mathbf{g}_{4}$ are the endpoints of the $k\textrm{th}$
ellipse, $\mathbf{g}_{2}$ is the endpoint of the $\left(k-1\right)\textrm{th}$
ellipse and $\mathbf{g}_{3}$ is the endpoint of the $\left(k+1\right)\textrm{th}$
ellipse. The centers are deleted if their distances to the common
vertices are smaller than $H$.

\begin{figure*}
\begin{centering}
\makebox[0.75\linewidth][c]{\subfigure[]{\includegraphics[width=0.2\paperwidth]{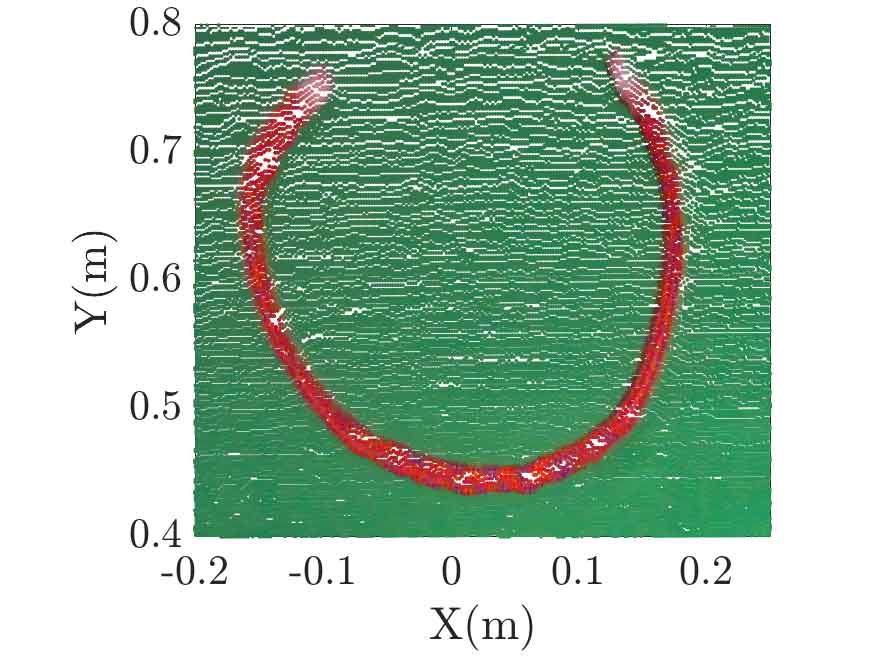}}\hspace{0.5em}\subfigure[]{\includegraphics[width=0.2\paperwidth]{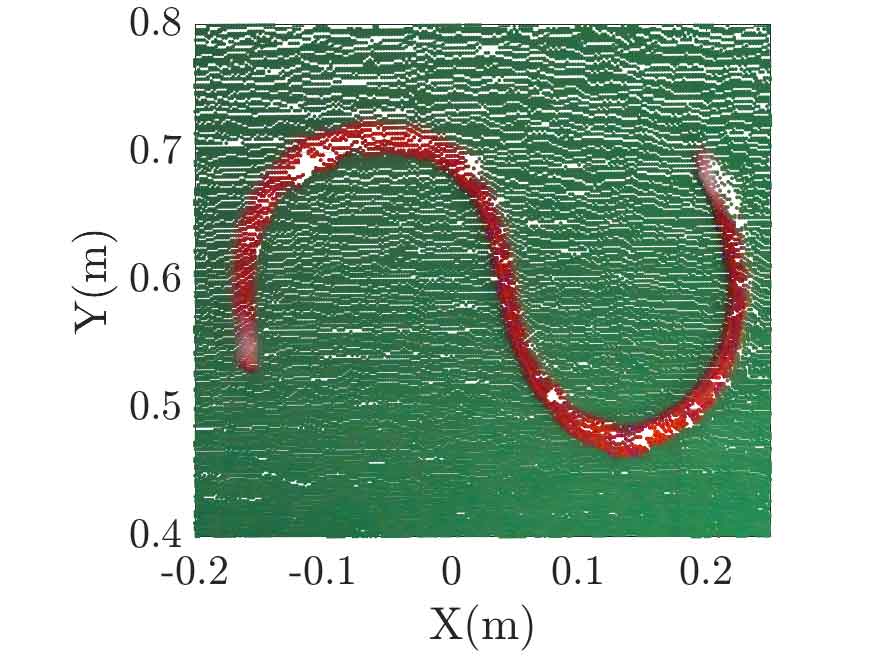}}\hspace{0.5em}\subfigure[]{\includegraphics[width=0.2\paperwidth]{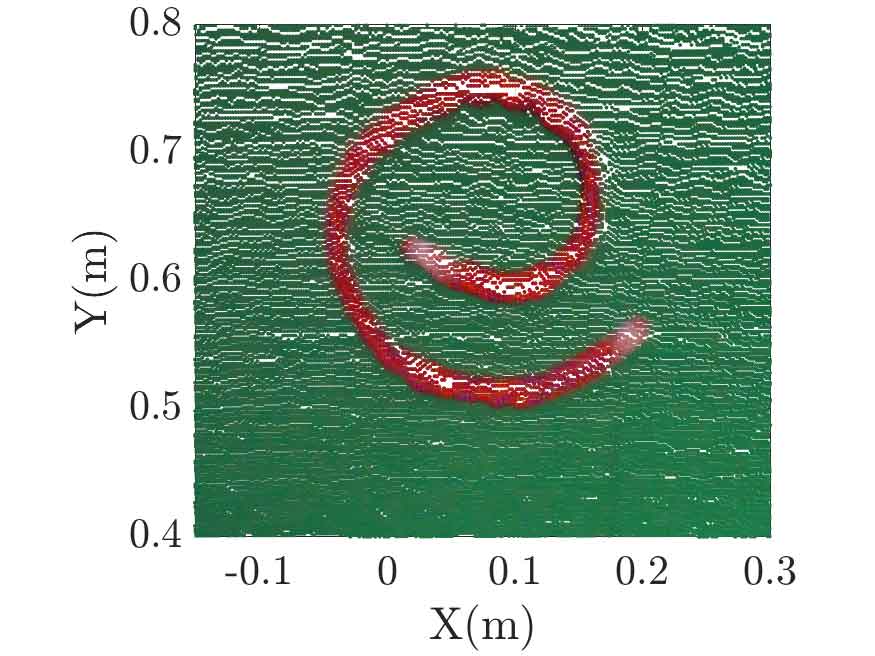}}}\\\vspace{-1em}
\par\end{centering}
\begin{centering}
\makebox[0.75\linewidth][c]{\subfigure[]{\includegraphics[width=0.2\paperwidth]{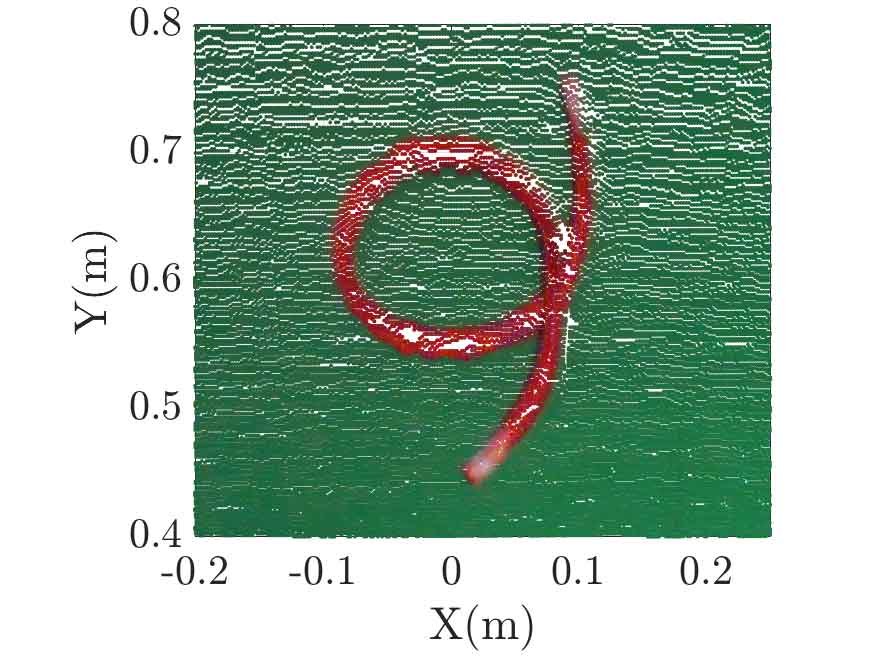}}\hspace{0.5em}\subfigure[]{\includegraphics[width=0.2\paperwidth]{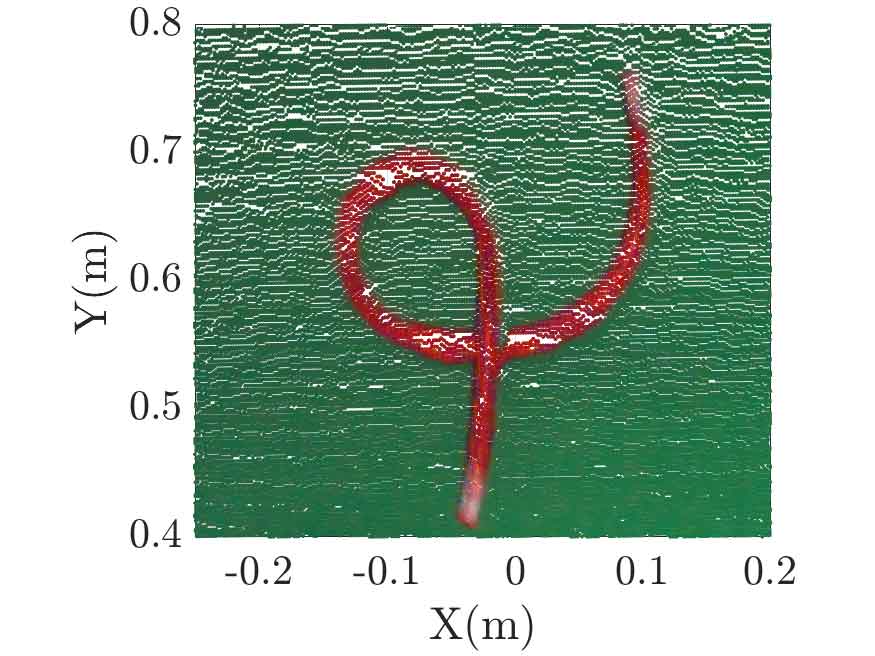}}\hspace{0.5em}\subfigure[]{\includegraphics[width=0.2\paperwidth]{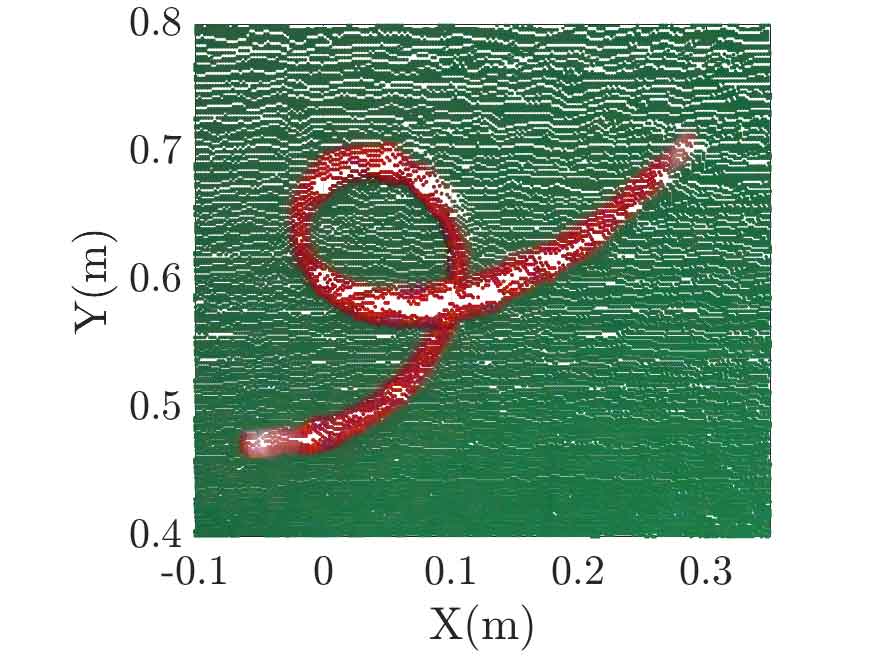}}}\\
\par\end{centering}
\centering{}\caption{Point clouds of the rope in different configurations obtained from
an RGB-D camera.\label{fig:Point-clouds-of-6-ropes}}
\end{figure*}

\section{Experimental Results\label{sec:Experimental-Results}}

In order to validate the proposed B-spline chained RMMs, a series
of experiments are performed to estimate the shape of the elongated
deformable objects. Intersection over union (IoU) is used as the metric
to evaluate the accuracy of shape estimation of the proposed algorithm.
The IoU is defined as the area of intersection of the estimated shape
and the true shape divided by the union of the two shapes \cite{zea2016tracking}
\begin{equation}
\mathbf{IoU}=\frac{area(\mathbf{\theta^{*}})\cap area(\mathbf{\hat{\theta}})}{area(\mathbf{\theta^{*}})\cup area(\mathbf{\hat{\theta}})}
\end{equation}
where $\mathbf{\hat{\theta}}$ is the true shape parameters and $\mathbf{\theta^{*}}$
is the estimated shape parameters. IoU is between $0$ and $1$, where
the value $1$ corresponds to a perfect match between the estimated
area and the ground-truth. Since the ground-truth of the position
and the shape of the elongated deformable object is difficult to obtain,
the measurements from the RGB-D camera are used as the ground-truth
\cite{zea2016tracking}. The measurement noise is neglected because
it is small compared to the area of the elongated deformable object.
The ground-truth is constructed by creating a $1\textrm{px}\times1\textrm{px}$
rectangle at each measurement point and taking the union of all the
rectangles \cite{zea2016tracking}. The dilation and erosion methods
are applied to ensure that the pixels are fully connected while preserving
the boundary of the target.

\begin{figure*}[t]
\begin{centering}
\makebox[0.7\linewidth][c]{\subfigure[]{\includegraphics[width=0.3\paperwidth]{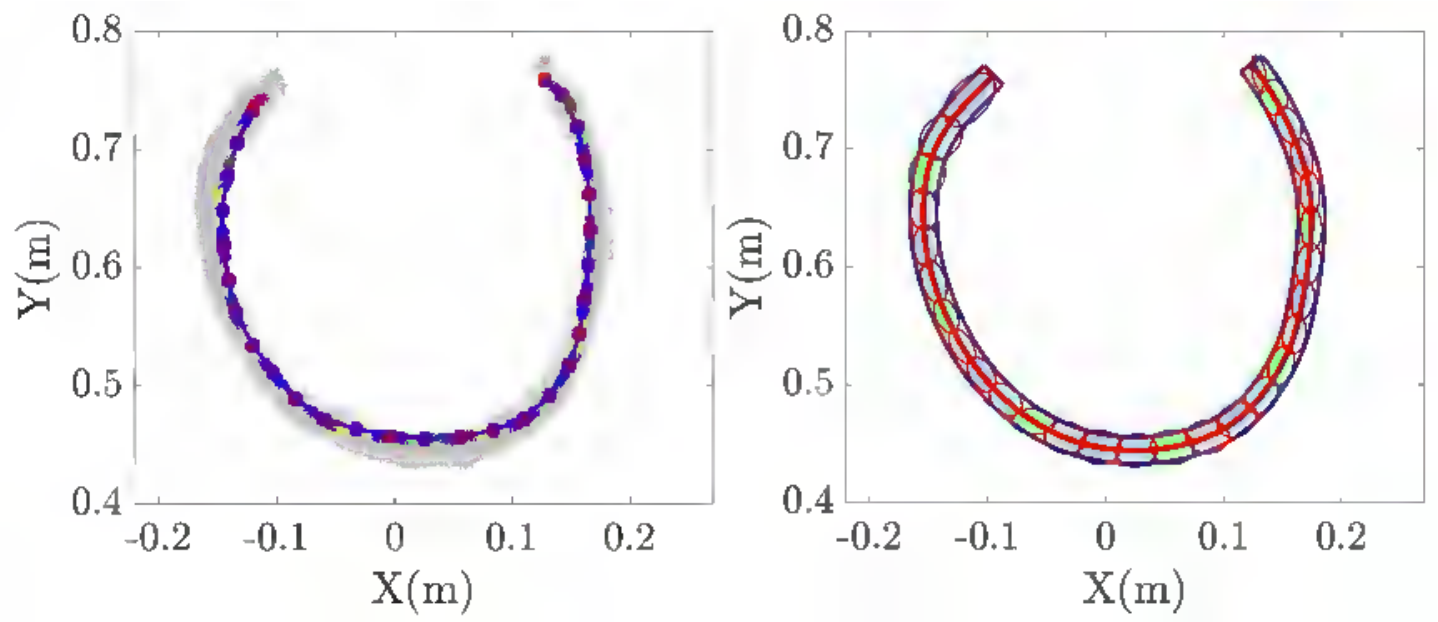}}\hspace{1em}\subfigure[]{\includegraphics[width=0.3\paperwidth]{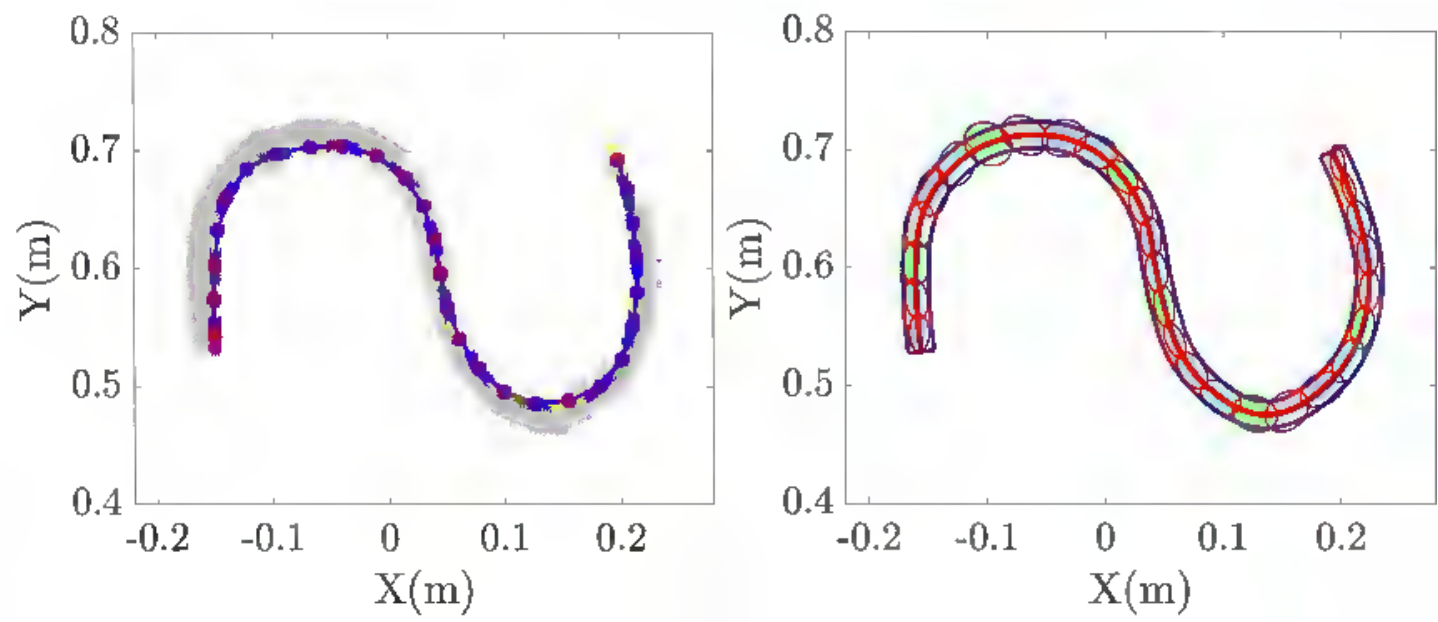}}}\\
\par\end{centering}
\begin{centering}
\makebox[0.7\linewidth][c]{\subfigure[]{\includegraphics[width=0.3\paperwidth]{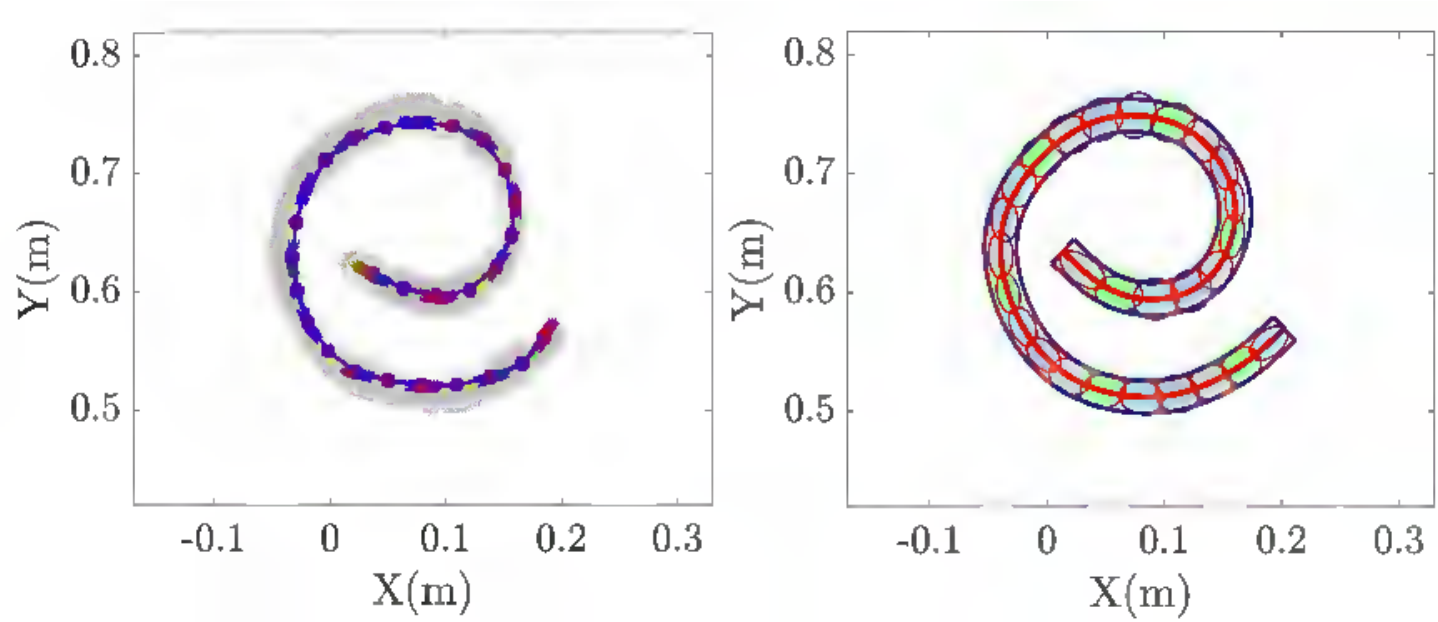}}\hspace{1em}\subfigure[]{\includegraphics[width=0.3\paperwidth]{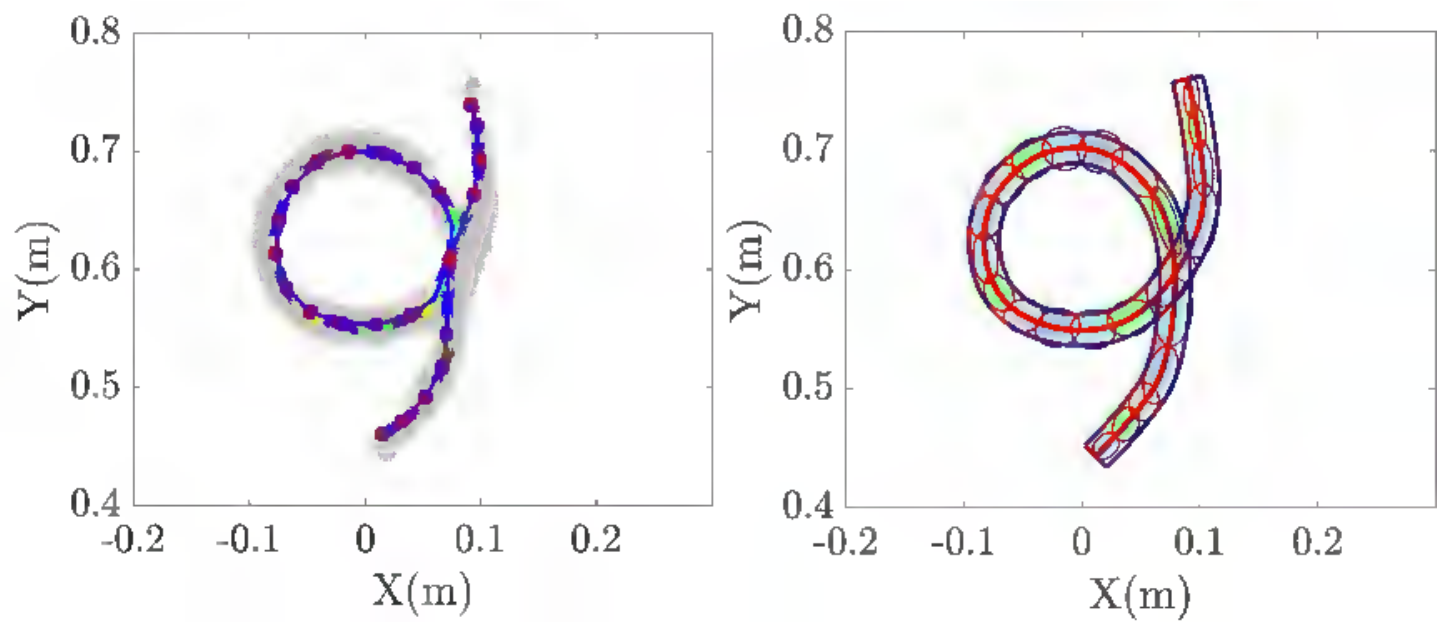}}}\\
\par\end{centering}
\begin{centering}
\makebox[0.7\linewidth][c]{\subfigure[]{\includegraphics[width=0.3\paperwidth]{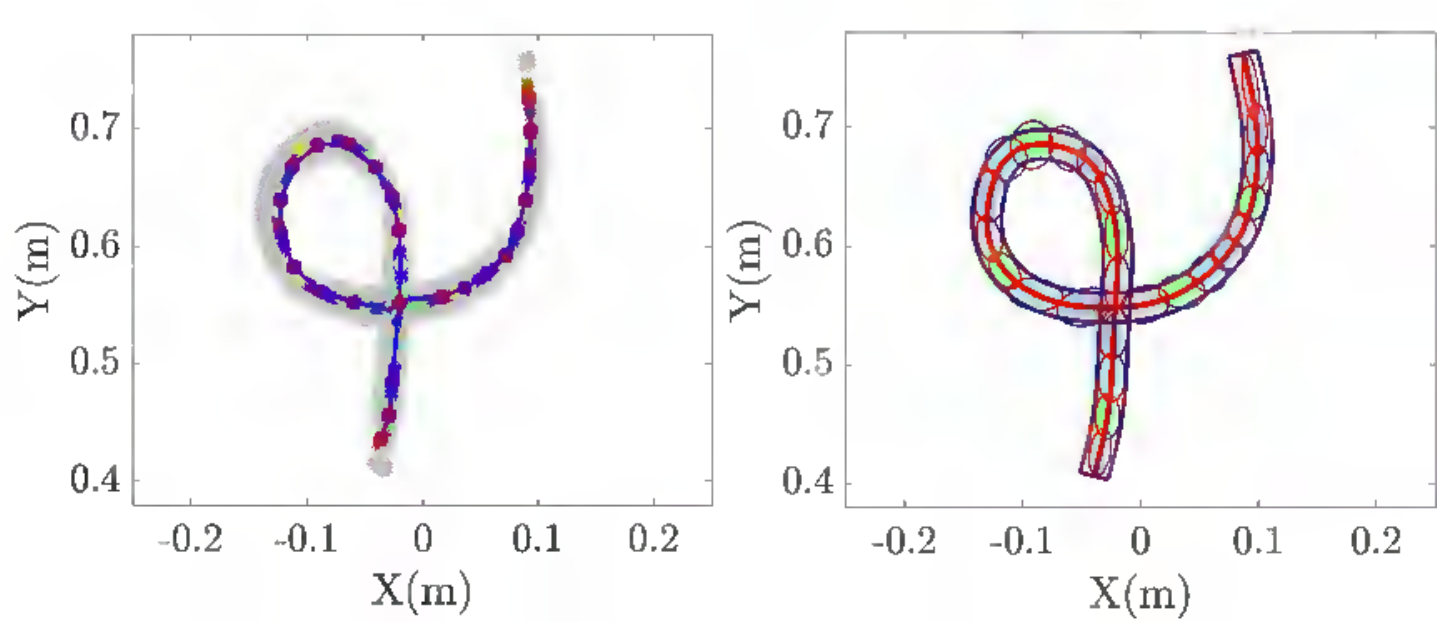}}\hspace{1em}\subfigure[]{\includegraphics[width=0.3\paperwidth]{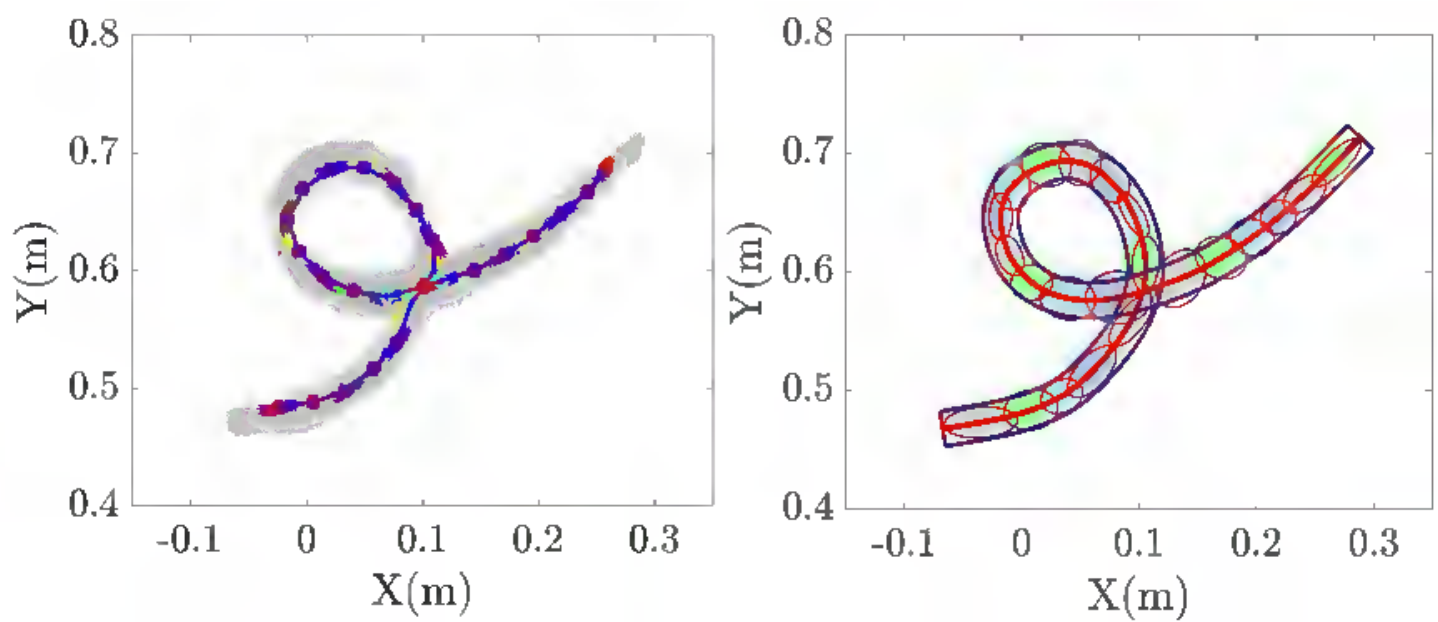}}}\\
\par\end{centering}
\centering{}\caption{Shape estimation results on the rope in 6 different configurations.
For each sub-figure (a)-(f), the left plot is the initialization result
and the right plot is the estimated shape of the rope.\label{fig:Shape-estimation-results}}
\end{figure*}

\subsection{Shape Estimation Results of The Static Rope}

A red nylon dock line with width of $15.8\textrm{mm}$ and length
of $930\textrm{mm}$ is used in the first part of the experiment.
The rope is manipulated to $6$ different shapes either with or without
intersection. A green cloth is used as the background. A Microsoft\textsuperscript{\textregistered}
Kinect camera is used to obtain a point cloud of the rope. Plane fitting
method and clustering algorithms are applied to subtract the rope
point cloud from the background. The point clouds of the rope with
$6$ different configurations are shown in Fig. \ref{fig:Point-clouds-of-6-ropes}.

The shape estimation results on $6$ different configurations of the
rope are shown in Fig. \ref{fig:Shape-estimation-results}. Each sub-figure
includes two plots. The left plot is the initialization result. The
gray measurement points are the measurements from the rope. The colored
segments are the initial segments. The red dots are the centers of
each segment and the blue line segments are the initial graph of the
centers. The right plot is the estimated shape of the rope. The segmented
measurement points are denoted by different colors. The red ellipses
represent multiple RMMs and the red curve is the B-spline curve of
the centers of the ellipses. The offset curves in blue in Fig \ref{fig:Shape-estimation-results}
are created by shifting the estimated B-spline curve by half of the
estimated width of the rope. The $H$ value in (\ref{eq:segment-criteria})
in the initialization procedure is set to $30\textrm{mm}$. The degree
of the B-spline curve is set to $2$ and the number of the control
points is set to $13$ for all 6 configurations. The maximum iteration
number of the EM algorithm is set to $3$. 

\begin{table}
\centering{}\caption{The estimated width and IoU values of the rope in $6$ different configurations.\label{tab:The-estimated-thickness-iOu}}
\resizebox{\columnwidth}{!}{\centering%
\begin{tabular}{>{\centering}p{2cm}cccccc}
\toprule 
\centering{}Configuration & 1 & 2 & 3 & 4 & 5 & 6\tabularnewline
\midrule
\midrule 
\multirow{2}{2cm}{\centering{}Width (mm)} & \multirow{2}{*}{11.4 } & \multirow{2}{*}{11.2} & \multirow{2}{*}{13.7} & \multirow{2}{*}{13.0} & \multirow{2}{*}{12.3} & \multirow{2}{*}{14.9}\tabularnewline
 &  &  &  &  &  & \tabularnewline
\midrule 
\centering{}IoU & 0.752 & 0.738 & 0.731 & 0.732 & 0.736 & 0.715\tabularnewline
\midrule 
\centering{}IoU\cite{schulman2013tracking} & 0.683 & 0.654 & 0.632 & 0.603 & 0.642 & 0.520\tabularnewline
\bottomrule
\end{tabular}}
\end{table}

The estimated width of the rope and the IoU values are shown in Table \ref{tab:The-estimated-thickness-iOu}.
The estimated width is smaller than the true width (15.8mm) and the
average IoU value over 6 configurations is $0.734$. Because the surface
of the rope is not flat, the detection of the edge points of the rope
using Kinect sensor is hard. This causes the estimates of the width
of the rope to be smaller than the ground-truth. Algorithm in \cite{schulman2013tracking} is also applied to estimate
the shape of the rope in $6$ configurations, of which the code is
publicly available (\url{http://rll.berkeley.edu/tracking/}). The
algorithm in \cite{schulman2013tracking} models a virtual rope as
a set of serial-linked capsules in a simulation. The radius of the capsules
is set as 7.9mm and other parameters are set to the default settings.
The shape estimations of the proposed algorithm are more accurate
compared with the algorithm in \cite{schulman2013tracking}, based
on the IoU values shown in Table \ref{tab:The-estimated-thickness-iOu}. 

The experiments are performed by measuring the elongated deformable
object on a tabletop with an RGB-D sensor. The sensor only detects
the top surface of the elongated deformable object closest to the
sensor. Thus, the two dimensional shape is estimated by the proposed
algorithm and the third dimension is determined by the tabletop. However,
the proposed model can be extended to estimate the shape of the rope
in three dimensional space. The control points of the B-spline curve
in (\ref{eq:B-spline}) can be extended into three dimensions as $\mathbf{b}_{i}\in\mathbb{R}^{3\times1}$
\cite{de1978practical}. RMM can also be extended into three dimensions
to represent an ellipsoid \cite{feldmann2011tracking}.

\begin{table}
\centering{}\caption{Typical execution time for initialization and EM algorithms with the
rope in $6$ different configurations.\label{tab:execution time}}
\resizebox{\columnwidth}{!}{\centering%
\begin{tabular}{>{\centering}p{3cm}cccccc}
\toprule 
\centering{}Configuration & 1 & 2 & 3 & 4 & 5 & 6\tabularnewline
\midrule
\midrule 
\centering{}Initialization (sec) & 0.195 & 0.196 & 0.199 & 0.221 & 0.212 & 0.215\tabularnewline
\midrule 
\centering{}EM (sec) & 0.296 & 0.291 & 0.283 & 0.253 & 0.248 & 0.252\tabularnewline
\bottomrule
\end{tabular}}
\end{table}

\begin{figure*}[t]
\begin{centering}
\makebox[0.75\linewidth][c]{\subfigure[Rope1]{\includegraphics[width=0.15\paperwidth]{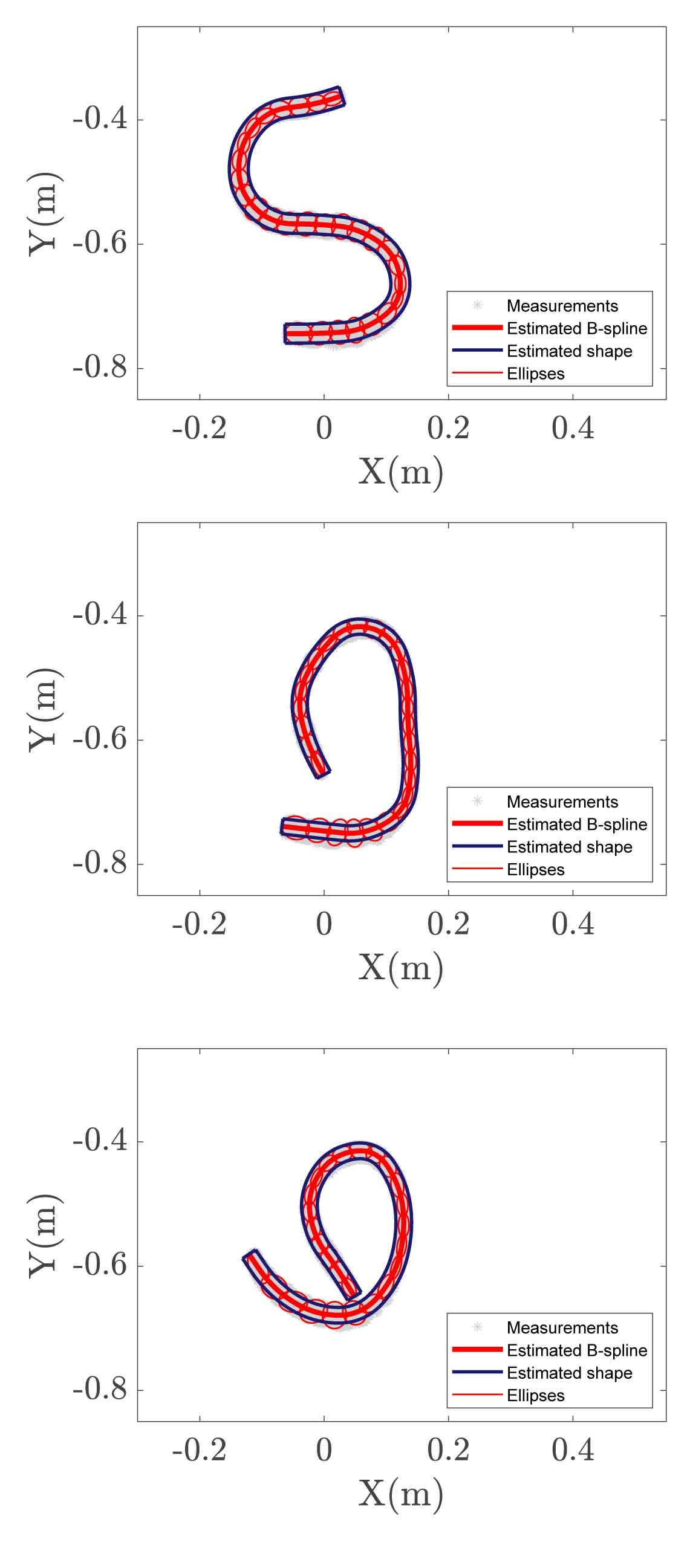}}\hspace{0.5em}\subfigure[Rope2]{\includegraphics[width=0.15\paperwidth]{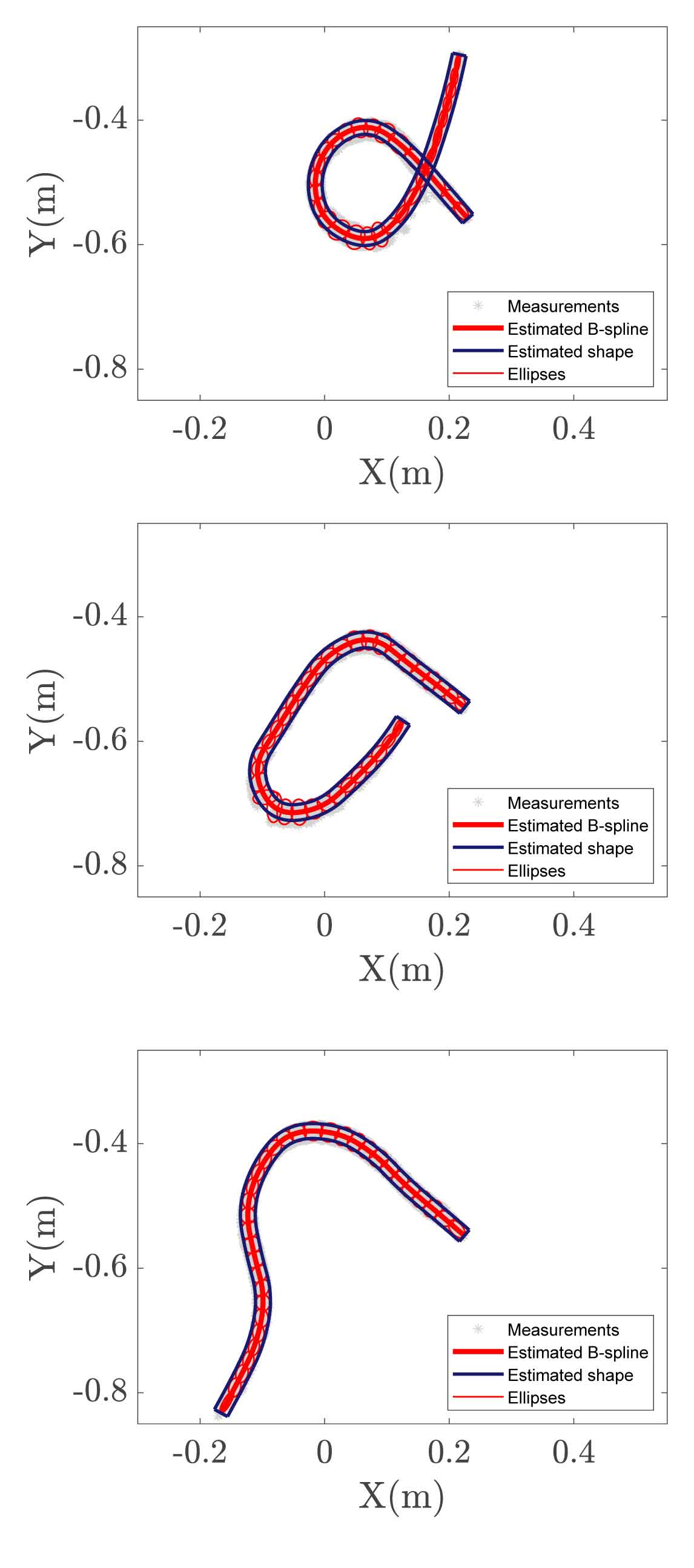}}\hspace{0.5em}\subfigure[Tube1]{\includegraphics[width=0.15\paperwidth]{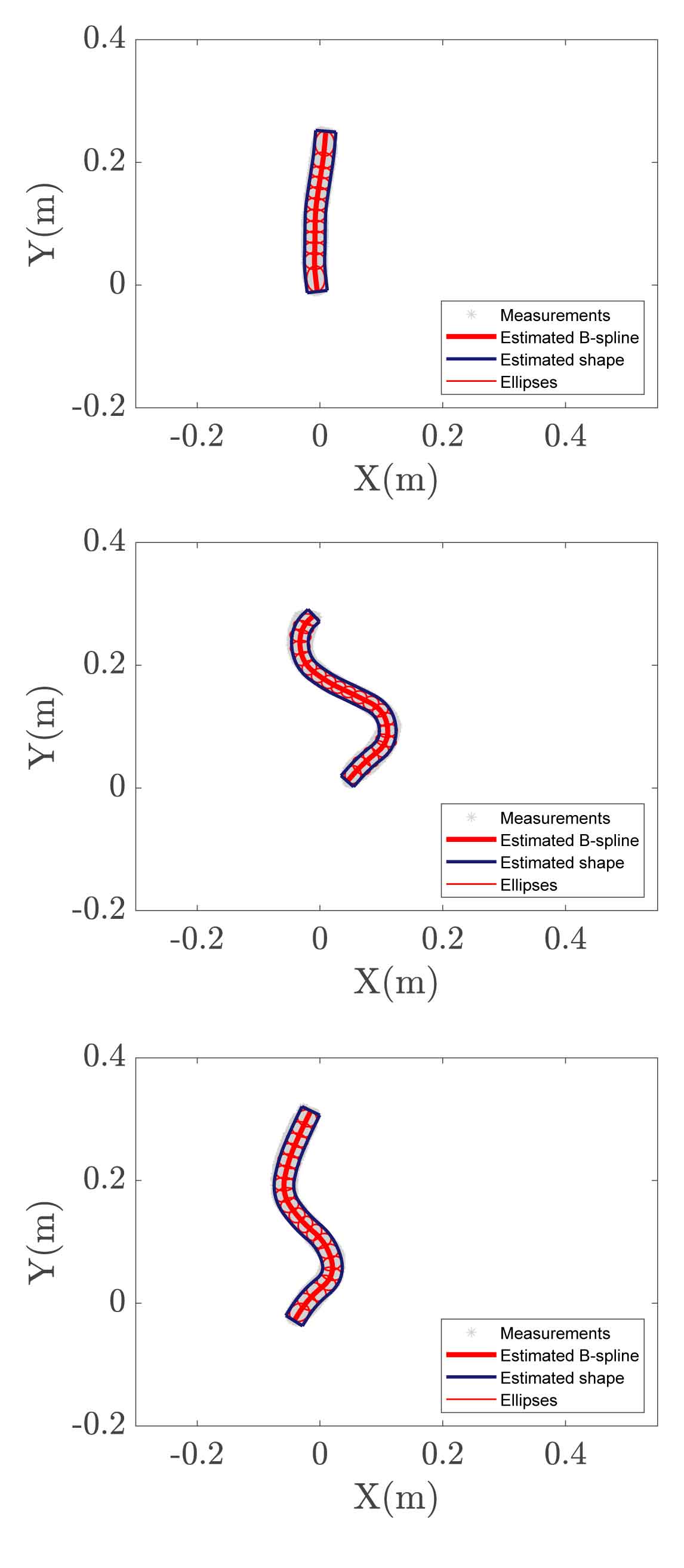}}\hspace{0.5em}\subfigure[Tube2]{\includegraphics[width=0.15\paperwidth]{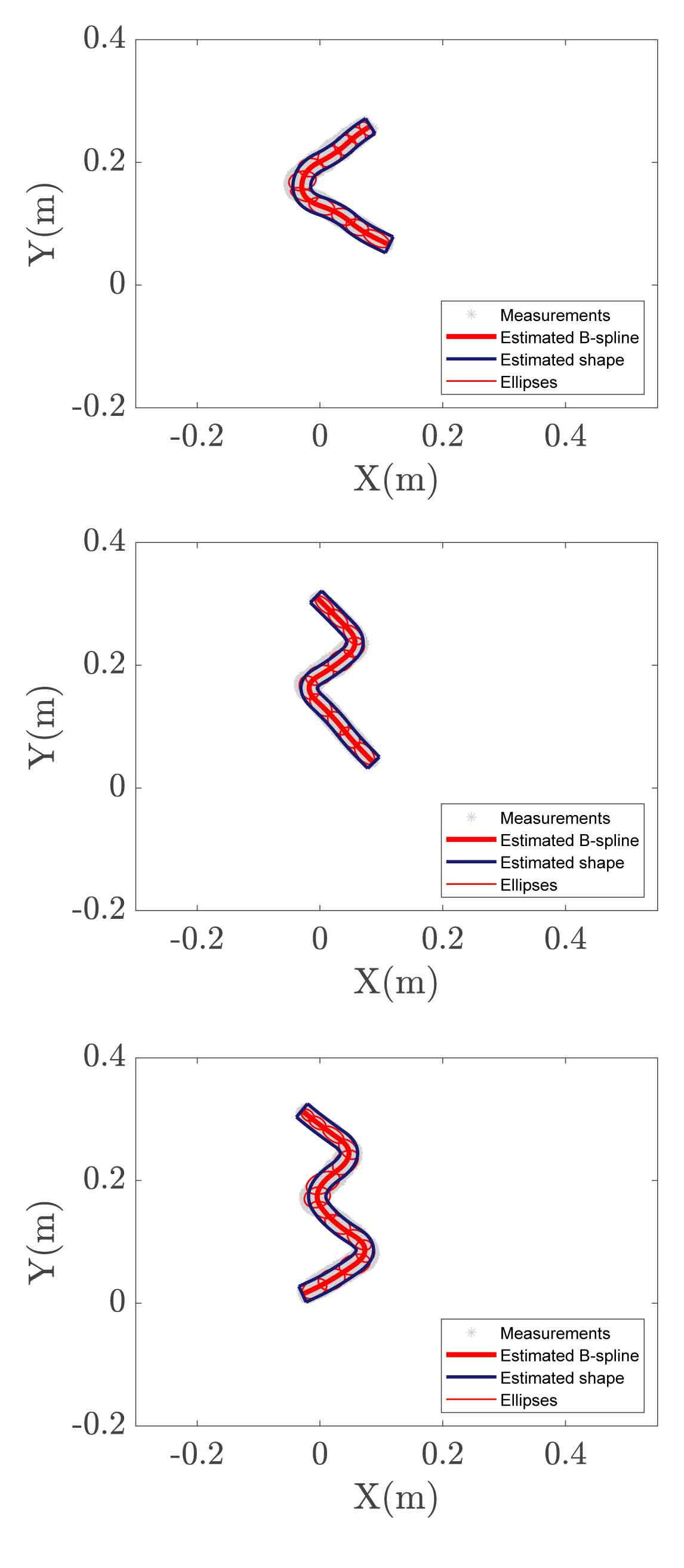}}\hspace{0.5em}\subfigure[Assembling]{\includegraphics[width=0.15\paperwidth]{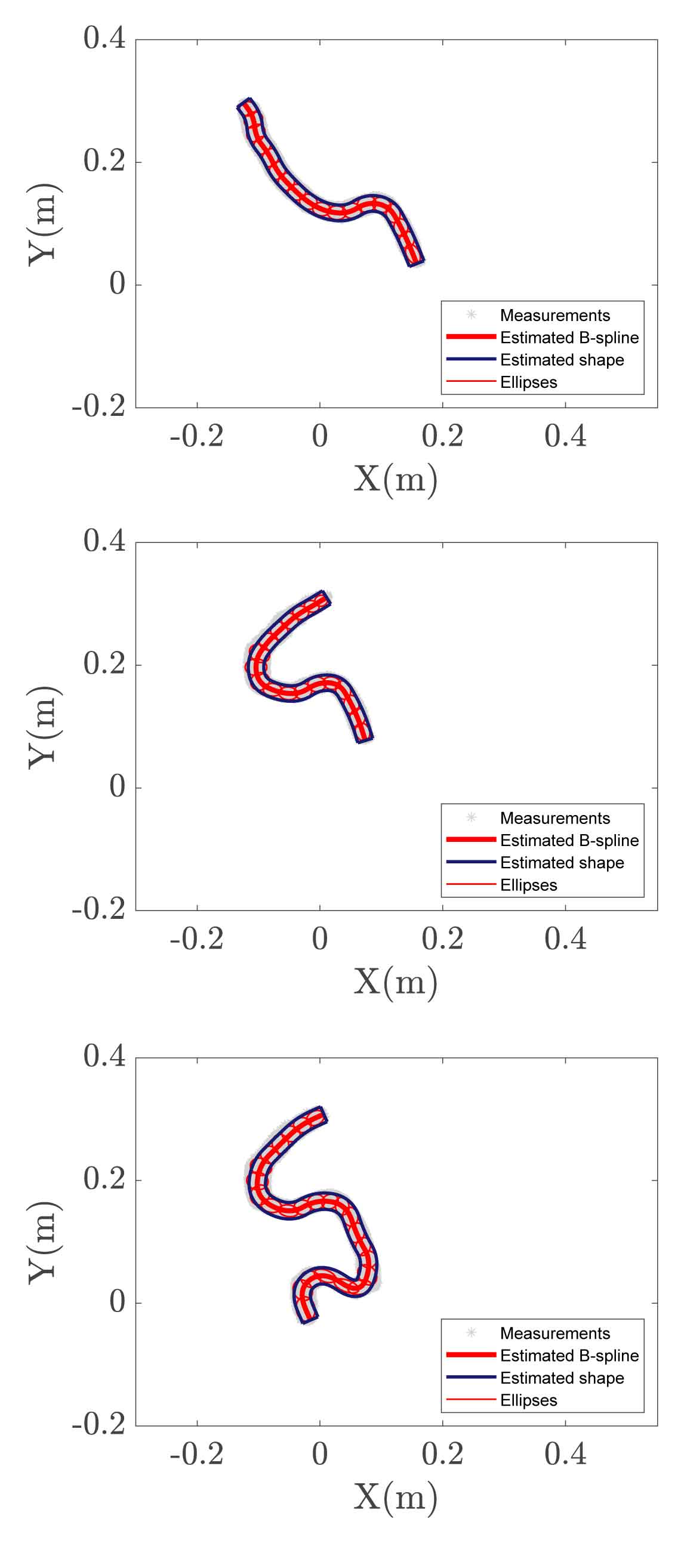}}}\\
\par\end{centering}
\centering{}\caption{Shape estimation results on the rope and the plastic tube during manipulations.
For each sub-figure (a)-(e), the $1^{st}$ frame, the $25^{th}$ frame
and the $50^{th}$ frame are shown from top to bottom.\label{fig:videos}}
\end{figure*}

The codes of the initialization and EM algorithms were run in MATLAB\textsuperscript{\textregistered}
R2019b on a Windows 10 PC with Intel\textsuperscript{\textregistered}
$\mathrm{i7-9700k@3.60GHZ}$ processor and $\mathrm{32.0GB}$ of RAM.
The typical execution time for initialization and EM algorithms are
shown in Table \ref{tab:execution time}. The most time consuming
parts of the initialization stage are building the EMST and the calculation
of the distance $D_{h}$ in (\ref{eq:dsitance}). The time complexity
of building the EMST is $\mathcal{O}(\left|\mathrm{\mathcal{E}}\right|\mathrm{log}\left|\mathcal{V}\right|)$,
where $\left|\mathrm{\mathcal{E}}\right|$ and $\left|\mathcal{V}\right|$
are the number of edges and vertices in graph $T=(\mathcal{V},\mathcal{E})$
\cite{sedgewick2011algorithms}. The calculation of distance $D_{h}$
in (\ref{eq:dsitance}) requires $\mathcal{O}(\left|\mathcal{V}_{p}\right|\left|\mathcal{C}_{h}\right|)$
operations, where $\left|\mathcal{V}_{p}\right|$ is the number of
vertices from $\mathcal{V}_{p}$ and $\left|\mathcal{C}_{h}\right|$
is the number of vertices from $\mathcal{C}_{h}$. The time complexity
of the EM algorithm is determined by the expectation stage, which
assigns the measurements into different ellipses. The time complexity
is $\mathcal{O}(KN_{r}),$ where $N_{r}$ is the number of the measurements
$\mathbf{Z}=\{\mathbf{z}_{r}\}_{r=1}^{N_{r}}$ and $K$ is the number
of clusters. In order to reduce the execution time for the shape estimation of the elongated deformable
object, parallel computing can be used or the number of measurements
can be uniformly down sampled. 

\begin{table}
\centering{}\caption{The average execution times and average IoU values of the rope and
the plastic tube during manipulations.\label{tab:video}}
\resizebox{\columnwidth}{!}{\centering%
\begin{tabular}{>{\centering}p{2.3cm}ccccc}
\toprule 
\centering{}Video name & Rope1 & Rope2 & Tube1 & Tube2 & Assembling\tabularnewline
\midrule
\midrule 
shape change & $\surd$ & $\surd$ & $\surd$ & $\surd$ & $\surd$\tabularnewline
\midrule 
length change &  &  & $\surd$ & $\surd$ & $\surd$\tabularnewline
\midrule 
execution time

(sec/frame) & 0.343 & 0.422 & 0.450 & 0.445 & 0.441\tabularnewline
\midrule 
\centering{}IoU & 0.727 & 0.709 & 0.799 & 0.768 & 0.819\tabularnewline
\bottomrule
\end{tabular}}
\end{table}

\subsection{Shape Estimation Results During Manipulations and Assembly}

The previous experiments show the shape estimation of the rope in
$6$ different configurations, based on the measurements from one
frame. The proposed EM algorithm is also used for shape estimation of the elongated
deformable object over multiple frames. The IoU value between the
estimated shape of the previous time step and the current measurements
is calculated for each frame. If the calculated IoU value is smaller
than a threshold, the initialization procedure is rerun in consideration
of the abrupt changes of the shape or the length of the elongated
object. 

Besides the red nylon dock line, a flexible red plastic tube with
modifiable shape and length is also used in the experiments. The elongated
deformable objects are manipulated in $5$ different scenarios. The
scenarios and the estimated results are recorded as videos (see the
multimedia attachment). Each video has $50$ frames of point cloud
measurements. The video `Rope1' shows the scenario that the rope is
manipulated from the shape `s' to the shape `9'. The video `Rope2'
demonstrates the scenario when the rope changes from one intersection
configuration to non-intersection configuration. The `Tube1' shows
that the plastic tube is stretched and squeezed which changes both
the shape and the length. The video `Tube2' illustrates that one part
of the plastic tube is stretched at a time and the red plastic tube
is manipulated from the shape `L' to the shape `M'. The last video
`Assembling' shows the situation when two plastic tubes are assembled
together as one tube. 

The typical frames and the estimation results
are shown in Fig. \ref{fig:videos}. The description of the videos
and the estimation results, including the average IoU value and the
average execution time over $50$ frames for each video, are shown
in Table. \ref{tab:video}. Algorithm in \cite{schulman2013tracking} is applied to estimate the
shape of the rope in `Rope1' and `Rope2' scenarios. The average IoU
values for `Rope1' and `Rope2' scenarios are $0.505$ and $0.608$
separately. The proposed algorithm achieves better accuracy in terms
of IoU. Because the Algorithm in \cite{schulman2013tracking} uses
linked rigid objects as the simulation model of the rope, it cannot
work for the scenarios (e.g. `Tube1', `Tube2' and `Assembling') when
the elongated object are changing both length and shape during the
manipulations.  

\section{Conclusions and Future Work\label{sec:Conclusion}}

To localize the elongated deformable object and to estimate its shape,
a B-spline chained multiple RMMs representation and its corresponding
EM algorithm are developed in this paper. Based on the sparse measurements
from an RGB-D camera, the proposed algorithm approximates the elongated
deformable object as a set of chained ellipses by using a B-spline
curve. Each ellipse is represented by an RMM, of which the center
represents the location and the covariance matrix determines the shape
of the ellipse. All the centers are enforced to be located on a B-spline
curve, which represents the shape of the elongated deformable object.
The EM algorithm and its initialization method are presented to estimate
the control points of the B-spline curve as well as the RMMs. The
performance of the proposed shape estimation algorithm is evaluated
using real measurements of a red dock line in 6 different configurations.
The proposed algorithm is also used to estimate the shapes in scenarios
such as the continuous manipulation and the assembly of elongated
deformable objects. From the experimental results, it can be concluded
that the B-spline curve chained RMM algorithm is capable of estimating
the shape of the elongated deformable object configured as intersecting
and non-intersecting shapes. The case when the rope has more than
one intersection, has knots or is piled up will be studied in the
future work.

\bibliographystyle{IEEEtran}
\bibliography{elongatedobjecttracking}

\begin{thebibliography}{10}
\providecommand{\url}[1]{#1}
\csname url@samestyle\endcsname
\providecommand{\newblock}{\relax}
\providecommand{\bibinfo}[2]{#2}
\providecommand{\BIBentrySTDinterwordspacing}{\spaceskip=0pt\relax}
\providecommand{\BIBentryALTinterwordstretchfactor}{4}
\providecommand{\BIBentryALTinterwordspacing}{\spaceskip=\fontdimen2\font plus
\BIBentryALTinterwordstretchfactor\fontdimen3\font minus
  \fontdimen4\font\relax}
\providecommand{\BIBforeignlanguage}[2]{{%
\expandafter\ifx\csname l@#1\endcsname\relax
\typeout{** WARNING: IEEEtran.bst: No hyphenation pattern has been}%
\typeout{** loaded for the language `#1'. Using the pattern for}%
\typeout{** the default language instead.}%
\else
\language=\csname l@#1\endcsname
\fi
#2}}
\providecommand{\BIBdecl}{\relax}
\BIBdecl

\bibitem{shah2018planning}
A.~Shah, L.~Blumberg, and J.~Shah, ``Planning for manipulation of interlinked
  deformable linear objects with applications to aircraft assembly,''
  \emph{IEEE Transactions on Automation Science and Engineering}, no.~99, pp.
  1--16, 2018.

\bibitem{zea2016tracking}
A.~Zea, F.~Faion, and U.~D. Hanebeck, ``Tracking elongated extended objects
  using splines,'' in \emph{19th International Conference on Information Fusion
  (FUSION)}, 2016, pp. 612--619.

\bibitem{sanchez2018robotic}
J.~Sanchez, J.-A. Corrales, B.-C. Bouzgarrou, and Y.~Mezouar, ``Robotic
  manipulation and sensing of deformable objects in domestic and industrial
  applications: a survey,'' \emph{The International Journal of Robotics
  Research}, vol.~37, no.~7, pp. 688--716, 2018.

\bibitem{jackson2017real}
R.~C. Jackson, R.~Yuan, D.-L. Chow, W.~S. Newman, and M.~C.
  {\c{C}}avu{\c{s}}o{\u{g}}lu, ``Real-time visual tracking of dynamic surgical
  suture threads,'' \emph{IEEE Transactions on Automation science and
  Engineering}, vol.~15, no.~3, pp. 1078--1090, 2017.

\bibitem{tang2018track}
T.~Tang and M.~Tomizuka, ``Track deformable objects from point clouds with
  structure preserved registration,'' \emph{The International Journal of
  Robotics Research}, 2018.

\bibitem{schulman2013tracking}
J.~Schulman, A.~Lee, J.~Ho, and P.~Abbeel, ``Tracking deformable objects with
  point clouds,'' in \emph{IEEE International Conference on Robotics and
  Automation (ICRA)}, 2013, pp. 1130--1137.

\bibitem{petit2015real}
A.~Petit, V.~Lippiello, and B.~Siciliano, ``Real-time tracking of 3d elastic
  objects with an {RGB-D} sensor,'' in \emph{IEEE/RSJ International Conference
  on Intelligent Robots and Systems (IROS)}, 2015, pp. 3914--3921.

\bibitem{javdani2011modeling}
S.~Javdani, S.~Tandon, J.~Tang, J.~F. O'Brien, and P.~Abbeel, ``Modeling and
  perception of deformable one-dimensional objects,'' in \emph{IEEE
  International Conference on Robotics and Automation (ICRA)}, 2011, pp.
  1607--1614.

\bibitem{DeGregorio}
D.~De~Gregorio, G.~Palli, and L.~Di~Stefano, ``{Let's} take a walk on
  superpixels graphs: Deformable linear objects segmentation and model
  estimation,'' in \emph{Asian Conference on Computer Vision (ACCV)}, 2018, pp.
  662--677.

\bibitem{lui2013tangled}
W.~H. Lui and A.~Saxena, ``Tangled: Learning to untangle ropes with rgb-d
  perception,'' in \emph{IEEE/RSJ International Conference on Intelligent
  Robots and Systems (IROS)}, 2013, pp. 837--844.

\bibitem{matsuno2006manipulation}
T.~Matsuno and T.~Fukuda, ``Manipulation of flexible rope using topological
  model based on sensor information,'' in \emph{IEEE/RSJ International
  Conference on Intelligent Robots and Systems (IROS)}, 2006, pp. 2638--2643.

\bibitem{yao2017image}
G.~Yao and A.~Dani, ``Image moment-based random hypersurface model for extended
  object tracking,'' in \emph{20th International Conference on Information
  Fusion (Fusion)}, 2017, pp. 1--7.

\bibitem{yao2018image}
G.~Yao, K.~Hunte, and A.~Dani, ``Image moment-based object tracking and shape
  estimation for complex motions,'' in \emph{American Control Conference
  (ACC)}, 2018, pp. 5819--5824.

\bibitem{feldmann2011tracking}
M.~Feldmann, D.~Franken, and W.~Koch, ``Tracking of extended objects and group
  targets using random matrices,'' \emph{IEEE Transactions on Signal
  Processing}, vol.~59, no.~4, pp. 1409--1420, 2011.

\bibitem{de1978practical}
C.~De~Boor, C.~De~Boor, E.-U. Math{\'e}maticien, C.~De~Boor, and C.~De~Boor,
  \emph{A practical guide to splines}.\hskip 1em plus 0.5em minus 0.4em\relax
  springer-verlag New York, 1978, vol.~27.

\bibitem{lee1989choosing}
E.~T. Lee, ``Choosing nodes in parametric curve interpolation,''
  \emph{Computer-Aided Design}, vol.~21, no.~6, pp. 363--370, 1989.

\bibitem{bishop2006pattern}
C.~M. Bishop, \emph{Pattern recognition and machine learning}.\hskip 1em plus
  0.5em minus 0.4em\relax springer, 2006.

\bibitem{huang2013l1}
H.~Huang, S.~Wu, D.~Cohen-Or, M.~Gong, H.~Zhang, G.~Li, and B.~Chen,
  ``{L1-medial} skeleton of point cloud.'' \emph{ACM Trans. Graph.}, vol.~32,
  no.~4, pp. 65--1, 2013.

\bibitem{lee2000curve}
I.-K. Lee, ``Curve reconstruction from unorganized points,'' \emph{Computer
  aided geometric design}, vol.~17, no.~2, pp. 161--177, 2000.

\bibitem{sedgewick2011algorithms}
R.~Sedgewick and K.~Wayne, \emph{Algorithms}.\hskip 1em plus 0.5em minus
  0.4em\relax Addison-Wesley Professional, 2011.

\end{thebibliography}

\end{document}